\documentclass[10pt,journal,compsoc]{IEEEtran}
\ifCLASSOPTIONcompsoc
  \usepackage[nocompress]{cite}
\else
  \usepackage{cite}
\fi
\usepackage[ruled,vlined]{algorithm2e}
\usepackage{multirow}
\usepackage{subcaption}

\usepackage{amsmath}
\usepackage{amssymb}
\usepackage{booktabs}
\usepackage{amsthm}

\def\calP{{\cal P}}

\def\calN{{\cal N}}

\makeatletter
\def\hlinewd#1{%
\noalign{\ifnum0=`}\fi\hrule \@height #1 \futurelet
\reserved@a\@xhline}
\makeatother

\ifCLASSINFOpdf
   \usepackage[pdftex]{graphicx}
\else
\fi
\def\calP{{\cal P}}
\def\calN{{\cal N}}

\usepackage{tikz}

\usepackage{wrapfig}

\def\hatu{\hat{u}}
\def\hato{\hat{o}}
\def\hatv{\hat{v}}

\def\calP{{\cal P}}

\def\calN{{\cal N}}
\def\calO{{\cal O}}
\newcommand{\bx}{\mbox{\boldmath $x$}}
\newcommand{\by}{\mbox{\boldmath $y$}}
\newcommand{\vbar}{\:|\:}

\begin{document}
\title{Occlusion-Ordered Semantic Instance Segmentation}

\author{{ Soroosh Baselizadeh, Cheuk-To Yu, Olga Veksler}
        and~Yuri~Boykov
\IEEEcompsocitemizethanks{\IEEEcompsocthanksitem  This work was performed while S. Baselizadeh was with the School of Computer Science, University of Waterloo. CT Yu, O. Veksler and Y. Boykov are with the School of Computer Science, University of Waterloo, Canada.\protect\\
E-mail:  cs.uwaterloo.ca/~oveksler/, cs.uwaterloo.ca/~yboykov/
%\IEEEcompsocthanksitem J. Doe and J. Doe are with Anonymous University.
}
% <-this % stops an unwanted space
} %Soroosh commented below line
\IEEEtitleabstractindextext{
\begin{abstract}

 Standard semantic instance segmentation provides useful, but inherently 2D information from a single image. To enable 3D analysis, one usually integrates absolute monocular depth estimation with instance segmentation. However, monocular depth is a difficult task. Instead, we leverage a simpler single-image task, occlusion-based relative depth ordering, providing coarser but  useful 3D information. We show that relative depth ordering works more reliably from occlusions than from absolute depth. We propose to solve the joint task of relative depth ordering and segmentation of instances based on occlusions. We call this task Occlusion-Ordered Semantic Instance Segmentation (OOSIS). 
 We develop an approach to OOSIS that extracts instances and their occlusion order simultaneously from oriented occlusion boundaries and semantic segmentation. 
 Unlike popular detect-and-segment framework for instance segmentation, combining occlusion ordering with instance segmentation allows a simple and clean formulation of OOSIS as a labeling problem. 
 As a part of our solution for OOSIS, we develop a novel oriented occlusion boundaries approach that significantly outperforms prior work. We also develop a new joint OOSIS metric based both on instance mask accuracy  and  correctness of their occlusion order. We achieve  better performance than strong baselines on KINS and COCOA datasets.  
\end{abstract}

}

\maketitle

\IEEEdisplaynontitleabstractindextext

% \fi
\IEEEpeerreviewmaketitle

\IEEEraisesectionheading{\section{Introduction}
\label{sec:intro}}

\IEEEPARstart{T}{he} task of semantic instance segmentation~\cite{he2017maskrcnn} is to generate a mask for each object from a set of classes, together with the class label. Semantic instance segmentation is useful for 2D scene analysis, but lacks depth information required for 3D scene understanding. On the other hand, depth provides strong cues for semantic instances. This motivates joint estimation of depth and instances. 

{\bf Absolute vs relative depth:} To enable 3D analysis of a single image, monocular estimation of absolute depth~\cite{saxena2005learning,eigen2014depth,godard2017unsupervised} is frequently integrated with instance (or panoptic) segmentation~\cite{zhang2018joint,xu2018pad,gao2022panopticdepth,dahnert2021panoptic,liu2021towards}.
However, absolute monocular depth requires real-valued estimates on the pixel level, which is a difficult task. Furthermore, as depth estimation degrades with distance, the absolute depth of far enough objects cannot be determined reliably. 
Even for close but thin objects, absolute depth is unlikely to distinguish them. 
An extreme example is a thin cardboard sheet against a wall. However, this is not a problem for {\em relative depth} that can be estimated from occlusions in a single image. It works for thin objects and degrades in accuracy with distance more gracefully, see Sec.~\ref{sec:experiments}.

\begin{figure}[!]
    \centering
    \includegraphics[width=8.5cm]{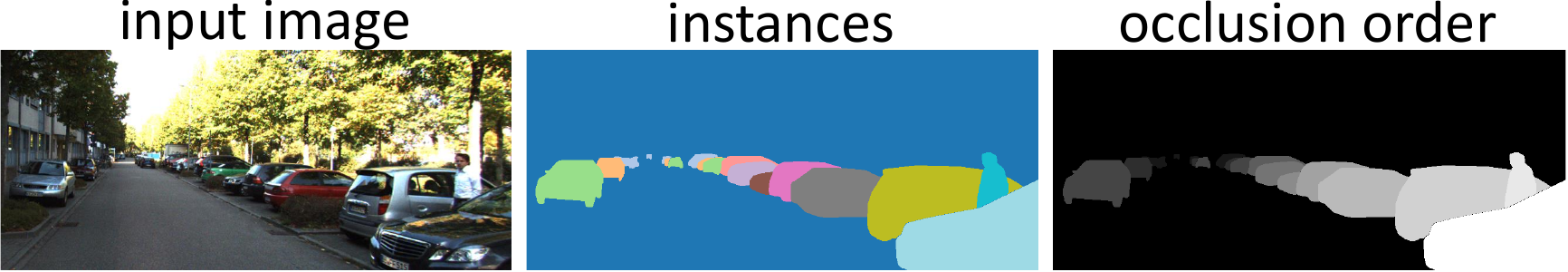}
    \caption{Illustrates occlusion-ordered semantic instance segmentation (OOSIS). Given an image, we output: (1) instances, (2) their occlusion order, visualized here as a relative depth map, i.e. an instance has a larger intensity than any neighboring instances it occludes.}
    \label{fig:teaser}
\end{figure}

Besides separating instances, occlusion is a powerful source of 3D information from a single image~\cite{hoiem2007recovering}, albeit for relative, not absolute depth. An occluding object is closer to the camera than its occludee.
Given  instances with pairwise
occlusion relations, we can recover the relative depth order between any instances connected by a monotonic chain: each object in the chain occludes the next object.
For example, in Fig.~\ref{fig:teaser},  eleven objects on the right are in a known relative depth order from pairwise occlusions.  
%For many scenes, this may establish the depth order between the majority of instances.
%Unlike monocular depth, occlusion cue is still useful for objects at large distances and for thin objects. 
Unlike absolute depth which requires real-valued labels, occlusion ordering needs only discrete labels, making it an easier task. This is confirmed by our experiments in Sec.~\ref{sec:experiments}, which show that relative depth ordering from occlusions is more reliable 
than from monocular depth. In human visual system, it is known that  occlusion provides depth ordering without attenuating with distance~\cite{cutting1995perceiving}.
Lastly, if not already collected as part of the original dataset, occlusion labels are easier to annotate than depth labels.
%Given a pair of objects, humans easily determine which one occludes the other, and there are datasets~\cite{Qi_2019_CVPR,zhu2017semantic} with such annotations.
%, usually used for amodal instance segmentation~\cite{Qi_2019_CVPR}. 

%One can also derive the relative depth order of instances from occlusions~\cite{yang2011layered,isola2013scene,tighe2014scene,zhang2015monocular}. 
%Interestingly,  humans are highly skilled at ordering objects by depth in images and even line drawings~\cite{cole2009well}. 

{\bf Problem formulation:} Motivated by the above, we propose to leverage occlusions as a single-image source of information for both instances and their relative depth, which allows 3D scene understanding.
We address \emph{Occlusion-Ordered Semantic Instance Segmentation} (OOSIS), which is the joint task of relative
% \begin{wrapfigure}{r}
% {0.5\textwidth}
%   \begin{center}
%     \includegraphics[width=6.5cm]{Fig/teaser1_cvpr.pdf}
%     \end{center}
%     \caption{Illustrates occlusion-ordered semantic instance segmentation (OOSIS). Given an image, we output: (1) instances, (2) their occlusion order, visualized here as a relative depth map, i.e. an instance has a larger intensity than instances it occludes.}
%     \label{fig:teaser}    
% \end{wrapfigure}
  occlusion-based depth ordering and segmentation of instances. 
We can visualize occlusion order using either a graph, where nodes are instances and directed edges are inserted for known adjacent occlusions, or a relative depth map, where an instance is assigned a larger intensity than any neighboring instance it occludes. Fig.~\ref{fig:teaser} is an illustration of OOSIS input and output.

 % Answering depth-related questions from standard semantic instance segmentation is not possible, at least not without further processing/learning. For example, a sofa can be to the left of a person in the image but it could be either behind or in front of the person in the corresponding 3D scene (from the camera's viewpoint). 

%The trade-off between monocular depth and occlusions is accuracy vs. generality. Monocular depth gives a more  detailed 3D scene geometry, but is more prone to errors (see~\cref{sec:experiments}). Based on occlusions, 
Unlike standard instance segmentation,
OOSIS can answer 3D geometry questions requiring depth ordering. For example, a sofa can be to the left of a person in the image as determined by standard instance segmentation, but it could be either behind or in front of the person in the corresponding 3D scene (from the camera's viewpoint). Standard instance segmentation cannot determine this depth relationship, but OOSIS can. OOSIS is useful in applications such as image captioning~\cite{karpathy2015deep}, question answering~\cite{malinowski2014multi}, 
 retrieval~\cite{johnson2015image}, 
 scene de-occlusion~\cite{zhan2020self}, anaglyph visualization~\cite{isola2013scene}  etc.

{\bf Our labeling approach:} We develop a method for OOSIS that performs instance segmentation and occlusion ordering simultaneously by formulating a Conditional Random Field (CRF) labeling problem~\cite{BVZ:PAMI01}. 
%We label pixels with their occlusion order. 
As it is difficult to come up with meaningful labels, most state-of-the-art  instance segmentation methods do not assign labels to instances directly, but are based on complex detect-and-segment frameworks. In contrast, combining occlusion ordering with instance segmentation allows a simple formulation of OOSIS as a clean labeling task where labels represent the occlusion order. Thus occlusion is a crucial component for forming instances. 
Another advantage of our approach is {\em unambiguous} segmentation by design,
i.e. each pixel belongs to at most one instance.

Our CRF model is based on an output of  CNN predicting oriented occlusion boundaries and semantic segmentation. We design a novel model for oriented occlusion boundaries, significantly outperforming prior work~\cite{wang2016doc,lu2019occlusion,wang2019doobnet,qiu2020pixel}.

Despite the existence of datasets annotated with instances and occlusions~\cite{Qi_2019_CVPR,zhu2017semantic}, our work is the first to address OOSIS with deep learning. The occlusion datasets are currently used for amodal instance segmentation~\cite{Qi_2019_CVPR}, or for improving standard instance segmentation~\cite{yuan2021robust,ke2023occlusion}, but not for OOSIS.
There are some works prior to deep learning for OOSIS~\cite{yang2011layered,isola2013scene,tighe2014scene}, however, they are based on simple heuristics, see Sec.~\ref{sec:related}. There are also methods which predict occlusion order given a pair of instance masks~\cite{zhu2017semantic,lee2022instance}. However, unlike us, these methods do not perform instance segmentation, they output {\emph{only}} an ordering of a {\emph{provided}} instant mask pair. We use such methods, combined with the state-of-the-art instance segmentation, for constructing strong comparison baselines for our work.

We also develop a new joint OOSIS metric which is based on both aspects of OOSIS: the quality of instance masks and the quality of the occlusion order.
We evaluate our approach on KINS~\cite{Qi_2019_CVPR} and COCOA~\cite{zhu2017semantic}.
We construct several  baselines combining prior works that separately address the two aspects of OOSIS 
(instance segmentation and occlusion-ordering).
We achieve better performance than the baselines in terms of the joint OOSIS metric, and other metrics.

 Our approach can also be regarded as a new unambiguous instance segmentation method, albeit requiring occlusion annotation for training. Our instance masks are more accurate than  the masks of state-of-the art unambiguous segmentation based on transformers, despite our architecture being less powerful. 
%Our method for OOSIS can be also regarded as a novel method for standard instance segmentation, but it requires occlusion annotated ground truth.

Our contributions are as follows: (a) the first deep learning based approach for OOSIS which cleanly formulates the task as a labeling problem; (b) state-of-the-art method for occlusion boundary detection; (c) a joint OOSIS metric for evaluation of OOSIS task; (d) focusing just on instance masks, our method is  a new clean labeling formulation for unambiguous instance segmentation.

\section{Related Work}
\label{sec:related}
{\bf{Semantic Instance Segmentation:}}
Multiple approaches to standard instance segmentation include  masks~\cite{he2017maskrcnn}, or   contours~\cite{xie2020polarmask,zhang2022e2ec}, or  bottom-up~\cite{liu2017sgn,newell2017associative}, etc.  There is also panoptic segmentation~\cite{Kirillov_2019_CVPR,cheng2021mask2former}, which produces unambiguous instance masks.
These works perform only one aspect (instance segmentation) of the joint OOSIS.

{{\bf Relative Depth Ordering of  Semantic Instances:}}
The most related to our work are~\cite{yang2011layered,isola2013scene,tighe2014scene}. They also propose to order  instances based on occlusions. However, they
predate deep learning and use simple hand-crafted occlusion strategies such as size, $y$-coordinate,  etc.
We take advantage of deep learning. 

The only deep learning work for instance segmentation based on ordering is in~\cite{zhang2015monocular}. 
 However, their ordering is based on depth, not occlusion, and we show in Sec.~\ref{sec:experiments} that occlusions are more reliable for ordering. Furthermore, they directly predict depth order to derive  instances, while we derive instances from detected occlusion edges. Detecting occlusion edges is a more local, and, therefore, an easier task compared to depth order prediction. Also, their formulation is based on patches, and the goal of their  CRF is to merge  patch-based predictions into a consistent global map. In our framework, CRF derives instances together with depth ordering simultaneously, based on oriented occlusion boundaries. Since the goals of CRFs are totally different, our energy functions are completely different. 
 %Because of direct depth order prediction of instances based on patches, they limit the number of possible depth layers to 6 in each patch. We do not have any such limitation. 
 Lastly, they handle only the car  class, whereas we handle any class.

{\bf{Integrated Depth and Segmentation:}} Also related is depth-aware semantic or panoptic segmentation~\cite{zhang2018joint,xu2018pad,gao2022panopticdepth}. Here the task is to produce segmentation together with the depth estimate from an  image. 
In~\cite{dahnert2021panoptic,liu2021towards}, they design depth-aware 3D panoptic segmentation. 
In contrast, in our OOSIS task, we leverage occlusion as a source for 3D information, which is a simpler single-image task and is more reliable for depth ordering than monocular depth  (see Sec.~\ref{sec:experiments}).  

{\bf{Occlusion Order Prediction:}}
There is prior work for occlusion order prediction~\cite{li2022distance,su20222,zhu2017semantic,lee2022instance}. These works address only one aspect of the joint OOSIS task, namely occlusion ordering. In particular, given for a pair of boxes~\cite{li2022distance,su20222} or instances~\cite{zhu2017semantic,lee2022instance}  they produce the occlusion order. For our baselines, we make use of~\cite{zhu2017semantic,lee2022instance} in combination with standard instance segmentation.
%often developed as a part of amodal instance segmentation in~\cite{zhu2017semantic,lee2022instance,zhan2020self}. The first two methods train on ground truth masks, the last method is self-supervised and less accurate. We make use of the first two (supervised) methods. 

{\bf Other Occlusion-related Work:} Some works improve instance or panoptic segmentation with occlusion-aware techniques~\cite{yuan2021robust,ke2023occlusion} or occlusion branches~\cite{Qi_2019_CVPR,lazarow2020learning}.
Our OOSIS task is different in essence, our goal is to solve joint  instance segmentation and occlusion ordering of instances, as opposed to improving standard instance segmentation.

{\bf{Oriented Occlusion Boundary Prediction:}}
Our OOSIS approach develops a method for oriented occlusion boundaries.  Prior work which treats the problem as regression, disentangles the boundary and orientation prediction~\cite{wang2016doc,lu2019occlusion,wang2019doobnet}. Prior work which treats the problem as classification does not disentangle the boundary and orientation prediction~\cite{qiu2020pixel}. Our model both treats the problem as classification and disentangles orientation and boundary prediction, outperforming the best prior work~\cite{qiu2020pixel}.

{\bf{Amodal Instance Segmentation:}}
Amodal instance segmentation methods~\cite{zhu2017semantic,Qi_2019_CVPR,zhan2020self} also use datasets with occlusion annotations. However, the goal of amodal segmentation is different from our OOSIS task: their aim is to extract both the visible and invisible part of a mask.

 \section{Our Approach to OOSIS}
% %%%%%% DIAGRAM OF OUR METHOD %%%%%%%
\begin{figure*}
  \centering
\includegraphics[width=0.7\textwidth]{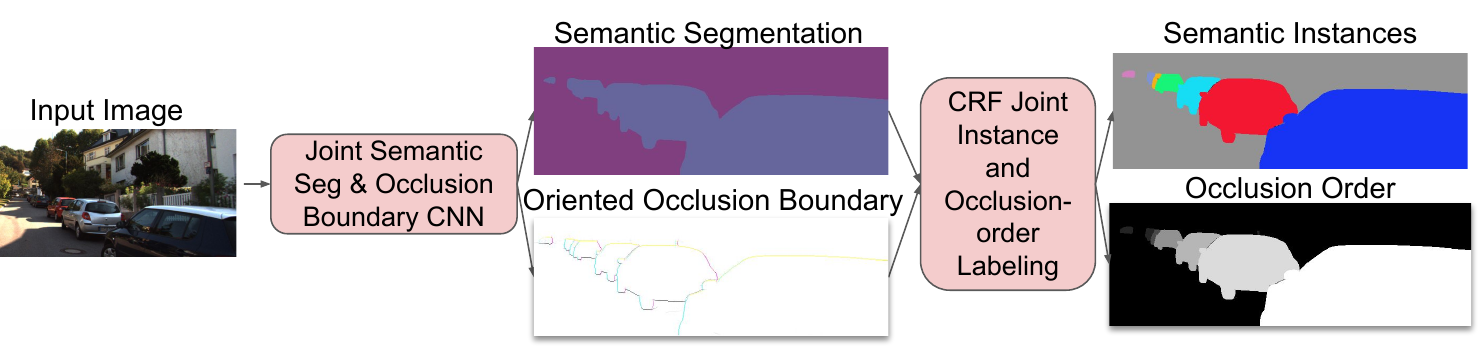}
  \caption{Overview of our approach.
    \label{fig:diagram}}
\end{figure*}

Our approach has two stages summarized in Fig.~\ref{fig:diagram}. The first stage (Sec.~\ref{sec:joint_sem_edge}) is based on deep learning and develops a novel model for simultaneously predicting semantic segmentation and oriented occlusion boundaries. The second stage (Sec. \ref{sec:instance_occlusion_labeling}) is based on discrete optimization, and  develops a novel CRF labeling approach for simultaneously inferring instances and their occlusion-order. 

  Unlike semantic segmentation, where it is easy to design labels (i.e. each class corresponds to a label), it is difficult to design meaningful labels for instance segmentation, i.e. labels that can be directly used for training.  Therefore, popular detect-and-segment frameworks~\cite{he2017maskrcnn}  do not assign labels to instances directly.
   In contrast, combining occlusion ordering with instance segmentation allows a simple formulation of OOSIS as a labeling problem.  Our  labels represent  occlusion-order: an instance with a larger label occludes an instance with a smaller label.   
Besides simplicity, our other advantage is unambiguous segmentation: each pixel is assigned to at most one instance. 

Before describing  our approach, we note that directly training CNN on occlusion-order labels fails, see Sec.~\ref{sec:naive_approach}. This is likely because occlusion-order labels do not have a stable meaning: instances with occlusion label $i$ across different scenes do not have much in common, in general. Our CRF framework  does not require label $i$ to mean the same thing across different images, label $i$ simply means that an instance with this label occludes any neighboring instance whose label is less than $i.$ 
%{\color{red} although it is correct, I think it might be a bit confusing as well, becuase in our labeling an instance with label${i}$ has certainly a neighbor with label ${i-1}$, feel free to keep it if you think it's clear enough.}

\subsection{Joint Semantic Segmentation and Oriented Occlusion Boundary Estimation } 
\label{sec:joint_sem_edge}

In the first stage, we design a joint CNN for semantic segmentation and oriented occlusion boundaries. A single CNN for both tasks saves computation. The architecture is modified PSPNet~\cite{zhao2017pspnet}, described in the Supplementary.

Our oriented boundary approach is new.
Prior work either disentangles boundary and orientation prediction~\cite{wang2016doc,lu2019occlusion,wang2019doobnet}, or treats the problem as classification, not regression~\cite{qiu2020pixel}. We do both, disentangle boundary and orientation, and treat the problem as classification,  outperforming the best prior work, see Sec.~\ref{sec:experiments}.

The occlusion orientation at a boundary pixel is described by the boundary's normal vector from occluder to occludee.
We introduce a random variable {\em oriented boundary} $\mathbb{O}_p \in \bar{D}$, where $\bar{D} = D\cup\{\emptyset\}$ and $D$ is the set of all possible orientations, e.g. discretized bins of  $360$-degree spectrum. Variable $\mathbb{O}_p$ has a dual purpose: 
besides indicating ``no boundary'' with $\emptyset$, in case of the boundary it indicates an orientation of its normal, an outcome in $D$. 

\begin{figure}
  \begin{center}
    \includegraphics[width=3cm]{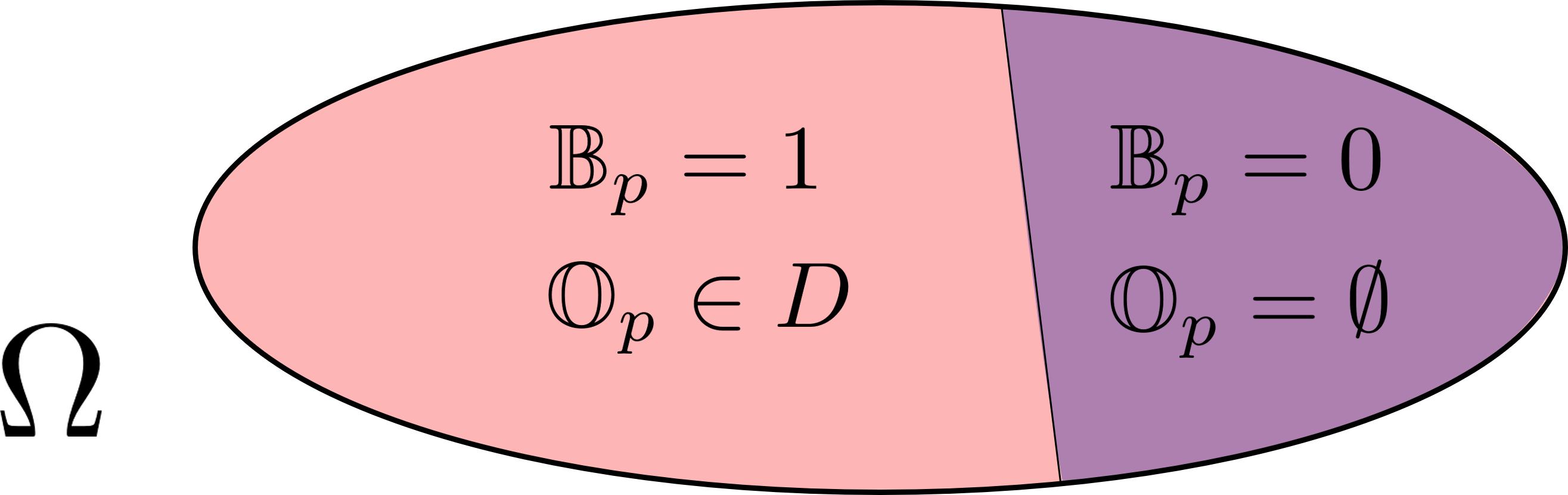}
    \end{center}
    \caption{Relationship between 
     ${\mathbb B}_p$ and ${\mathbb O}_p$. }
    \label{fig:relOB}    
\end{figure}

Instead of directly predicting oriented occlusion boundary $\Pr(\mathbb{O}_p)$ as in~\cite{qiu2020pixel}, to estimate $\Pr(\mathbb{O}_p)$, we disentangle boundary and orientation predictions by introducing a binary random variable $\mathbb{B}_p$ which indicates whether pixel $p$
  lies on an occlusion boundary. Fig.~\ref{fig:relOB} illustrates the relationship between ${\mathbb B}_p$ and ${\mathbb O}_p$, namely: ${\mathbb B}_p=1  \,\Leftrightarrow\,  {\mathbb O}_p\in D$ and
    ${\mathbb B}_p=0 \,\Leftrightarrow\, {\mathbb O}_p=\emptyset$. We predict ${\mathbb B}_p$ and
 $Pr(\mathbb{O}_p=d | \mathbb{B}_p = 1)$, the conditional distribution over $|D|$ orientations of the normal assuming $p$ is on a boundary. 
 Our modeling implies
\begin{align}
\Pr(\mathbb{O}_p=\emptyset) & = \Pr(\mathbb{B}_p=0)   \label{eq:O=empty} \\  
\Pr(\mathbb{O}_p = d) &= \Pr(\mathbb{O}_p=d\,|\,\mathbb{B}_p=1)  \Pr(\mathbb{B}_p=1)\;\forall d\in D   \label{eq:O=d}
\end{align}

To estimate probabilities we use two interdependent heads: binary 
boundary head $b_p$ and {\em conditional} orientation head $e_p$.
The head $b_p$ estimates $Pr(\mathbb{B}_p=1)$, while
$e_p$ estimates $Pr(\mathbb{O}_p=d | \mathbb{B}_p = 1)$. 
These interdependent heads replace classification heads in PSPNet~\cite{zhao2017pspnet}.

For the oriented occlusion boundaries, we need to obtain {\em oriented boundary} prediction $o_p:=Pr(\mathbb{O}_p)$, 
which is a distribution over $|D|+1$ values. 
 Eq.~(\ref{eq:O=empty}), (\ref{eq:O=d}) imply
a simple formal relation of this prediction to $b_p = \Pr(\mathbb{B}_p=1)$ and to
conditional orientation distribution $e_p=\Pr(\mathbb{O}_p=d | \mathbb{B}_p=1)$. 
Indeed, separating $|D|+1$ components of vector $o_p$ into two parts: $|D|$ values
corresponding to the probabilities $\Pr(\mathbb{O}_p=d)$ for $d\in D$ and 
an extra value corresponding to $\Pr(\mathbb{O}_p =\emptyset)$, equations Eq.~(\ref{eq:O=empty}), (\ref{eq:O=d}) give
\begin{equation}
    o_p := [b_p \odot e_p, 1-b_p],
\end{equation}
 where $[\cdot, \cdot]$ is concatenation and $\odot$ is element-wise multiplication.

We also introduce a random variable $\mathbb{S}_p$ to represent the semantic class of pixel $p$ and 
head  $s_p$ to estimate $Pr(\mathbb{S}_p)$, 
as in standard semantic segmentation.

% Ultimately, for every pixel $p$, we aim to predict $\Pr(\mathbb{S}_p)$, which represents the probability distribution over each semantic class, and $\Pr(\mathbb{O}_p)$, which represents the distribution of the normal orientation for a boundary pixel and also represents the probability of the pixel not being a boundary. 

To train  heads $s$, $b$, and $e$, we use the loss 
\begin{equation} \label{eq:our_3loss}
    \mathcal{L} = \sum_{p} CE(S_p | s_p) + CE_w(B_p | b_p) + B_p \cdot CE(E_p | e_p)
\end{equation}
where $CE$ is cross-entropy and $CE_w$ is {\em weighted} binary cross-entropy~\cite{sudre2017generalised}. 
The ground truth targets for $s_p$, $b_p$, and $e_p$ are denoted by $S_p$, $B_p$, and $E_p$, respectively. 
As clear from Eq.~(\ref{eq:our_3loss}), the conditional orientation head $e$ is trained only on pixels at 
the ground-truth occlusion boundary where $B_p=1$, which are the only pixels where the ground-truth normals $E_p$ 
are defined.  
In the Supplementary, we derive the loss in Eq.~(\ref{eq:our_3loss}) from cross-entropy between 
our model's final oriented occlusion boundary prediction $o_p$ and the observed ground-truth boundary normals.

Note that~\cite{wang2016doc} also separate boundary and orientation prediction. However, they predict orientation by regression on the angles, also trained on boundary points only. We use classification, 
which often works better than regression. We also provide a probabilistic derivation motivating 
our loss in Eq.~(\ref{eq:our_3loss}), see the Supplementary.

We use weighted cross-entropy $CE_w$ (with $w=0.9$) due to highly imbalanced ground truth 
as most pixels are non-boundary. 
We use 4 orientations ($|D|=4$) interpreted as left, right, top, and bottom neighbor occlusions, as in~\cite{qiu2020pixel}. 
For details on how we acquire the ground-truth orientations, see the Supplementary.

\begin{figure*}
  \centering
\includegraphics[width=0.9\textwidth]{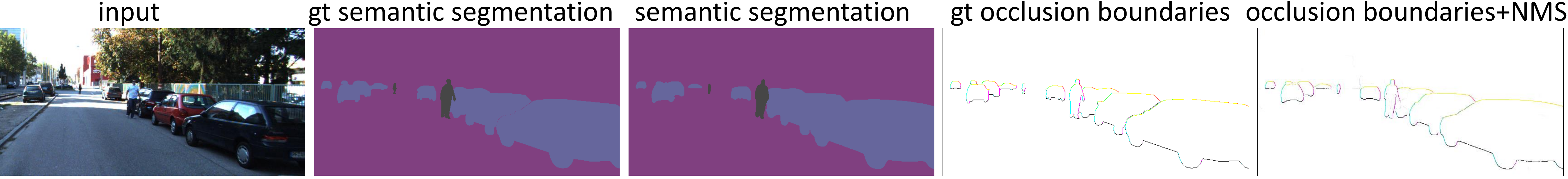}
  \caption{Illustration of our semantic segmentation and oriented occlusion boundaries. Images are: input, ground truth and our semantic segmentation, ground truth and our oriented occlusion boundaries after non-maximum suppression. The boundary color scheme: left-cyan, top-yellow, right-magenta, bottom-black. Best viewed zoomed in.}
   \label{fig:boundaries}
\end{figure*}

Fig.~\ref{fig:boundaries} illustrates our segmentation output $s_p$ and oriented occlusion boundary $b_p \odot e_p$ (part of $o_p$). To thin occlusion boundaries, we apply non-maximum suppression. We also discard boundaries below a threshold of $0.1$.

\vspace*{-1ex}
\subsection{CRF Occlusion-Order Labeling}
\label{sec:instance_occlusion_labeling}
In the second stage, we formulate OOSIS as a labeling problem in CRF discrete optimization framework~\cite{BVZ:PAMI01}, based on the semantic segmentation and oriented occlusion boundaries from the first stage.
To obtain instances with their occlusions jointly, we label pixels with their 
occlusion order. To enable formation of instances, we encourage a pixel to be assigned a larger label than the label of any neighboring pixel it is estimated to occlude according to the oriented boundaries. For spatially complete instances,
pixels which are not in the background according to semantic segmentation are encouraged to be inside some instance.

 We now describe  our formulation.
Let $x_p$ be the label assigned to pixel $p$.
The labels are non-negative integers and denote the occlusion-order.
 If $x_p=0$, then pixel $p$ is the background and does not occlude anything. Positive $x_p$ means that  pixel $p$ belongs to an instance. Given two neighboring pixels $p,q$, if $x_p > x_q$, then $p$ occludes $q$. A connected component of pixels with the same label forms an instance. Two distinct connected components with the same label are considered to be different instances. A drawback of our formulation is that we cannot have immediately adjacent instances with equal occlusion-order labels. However, immediately adjacent instances  are likely involved in an occlusion relation and have distinct occlusion-order which separates them, see, for example, ground truth in Fig.~\ref{fig:qual}. To determine the instance class, we take the majority vote on the semantic segmentation limited to the instance pixels.

Let $\calP$ be the set of  pixels, and  $\bx= (x_p\vbar p\in \calP)$ be  a labeling vector.
The energy  is defined for $\bx$ and  consists of unary and pairwise terms. 
 For each  $p$ there is a unary term $u_p(x_p)$. It is small if  label $x_p$ is likely for $p$ and large otherwise. We base unary terms on semantic segmentation, even though it outputs a distribution over  classes, not occlusion-orders. However, the occlusion-order $0$ is equivalent to the  semantic background class and is suitable for unary terms. 
  Let $\sigma_p$ be the probability for  $p$ to be the background according to semantic segmentation.  
  We set
\vspace*{-1ex}
\begin{equation}
    u_p(x_p) =   (1-\sigma_p) \cdot [x_p=0] +    \sigma_p \cdot [x_p \neq 0] ,
\end{equation}
where $[\cdot]$ is Iverson bracket, equal to $1$ if its argument is true and to $0$ otherwise. If $\sigma_p$ is large, the unary term for label $0$ is small, encouraging $p$  to be assigned to the background. If $\sigma_p$ is small, then $p$ is encouraged to be assigned to any non-zero label, without a particular preference among them. 
%as semantic segmentation is not informative about any occlusion-order label other than $0$. 

There are two types of pairwise terms: smoothness $v$ and occlusion $o$. 
The smoothness terms $v$ are modeled as in~\cite{BVZ:PAMI01}. They
encourage a spatially coherent labeling  by penalizing neighboring pixels that do not have the same label. Let $p,q$ be neighbors. The smoothness $v$ is defined as
\vspace*{-1ex}
\begin{equation}
    v(x_p,x_q)=[x_p\neq x_q].
\end{equation}

The occlusion terms $o$  are based on oriented occlusion boundaries,
which we store as a set of ordered pixel pairs: $\calO =\{ (p,q) \,|\, p \text{ occludes } q \}$. 
 Let $(p,q)\in\calO$, i.e.  $p$ occludes $q$. We assign a prohibitive cost if the label of $q$ is  larger than that of $p$, and  a negative cost if the label of $p$ is larger
 \vspace*{-1ex}
\begin{equation}
    o(x_p,x_q)=c_{\infty}\cdot [x_p<x_q]-[x_p>x_q],
\end{equation}
where $c_{\infty}$ is prohibitively large.
 Since we are minimizing the energy, a  negative cost is 'repulsive'~\cite{yu2001segmentation}, and we lower the energy if $p$ gets assigned a larger label than  $q$. Thus, 
 occlusion terms encourage  boundaries in a labeling $\bx$, facilitating creation of instances, unlike the smoothness terms, which discourage boundaries. But boundaries in $\bx$ are encouraged {\emph{only}} in places where we detect occlusion boundaries, which is important to maintain the  spatial  coherence of labeling.

\begin{figure*}[h]
  \centering
  \includegraphics[width=.9\textwidth]{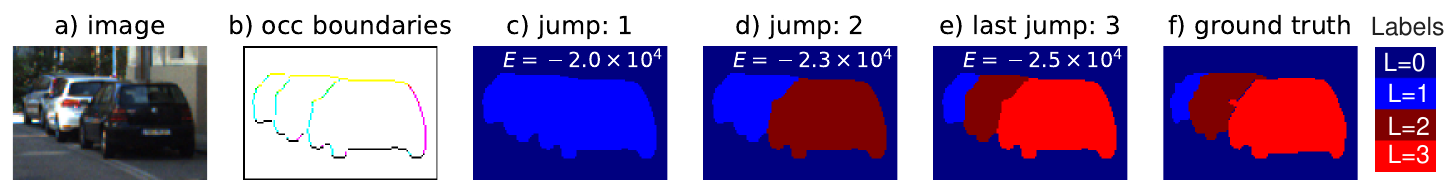}
  \caption{Jump move optimization.  Warmer colors correspond to larger labels. Energies are listed.}
   \label{fig:jumps}
\end{figure*}

The best labeling $\bx$ is found by minimizing the energy
%\vspace*{-1ex}
  \begin{equation} 
E(\bx)=  \sum_{p \in \calP}  u_p(x_p) + \lambda \sum_{(p,q)\in \calN} v(x_p,x_q) + \mu \sum_{(p,q)\in \calO}  o(x_p, x_q),
\label{eq:main-energy}
\end{equation}
where $\calN$ is the set of pairs on an 4-connected grid. 
We set $\lambda=20$ and $\mu=100.$

Fig.~\ref{fig:jumps} illustrates how  energy in Eq.~(\ref{eq:main-energy}) encourages instances to form along occlusion boundaries. 
% The energy decreases whenever for any $(p,q)\in\calO$, the label of $p$ is larger than the label of $q$, or, in other words, when the occluder pixel has a label larger than that of the pixel it occludes.
 The energy decreases when the occluder has a  label larger than the label of the occludee across occlusion boundaries. If this happens, we say that we ''take advantage'' of an occlusion boundary.
%Consider labelings  (c, d, e). 
Labeling in Fig.~\ref{fig:jumps}c has one instance which groups all cars together. It does not take advantage of the occlusion boundaries between the  cars. 
%These 'missed' occlusion boundaries can be used to lower the energy.
Labeling in Fig.~\ref{fig:jumps}d has two instances and takes advantage of some in-between the car boundaries. Labeling in Fig.~\ref{fig:jumps}e has the lowest energy because it takes advantage of all occlusion boundaries.

% To optimize CRF energies, the expansion algorithm~\cite{BVZ:PAMI01} is widely used.  We found that the jump move algorithm~\cite{Veksler:phd} works better, see Supplementary for an explanation. For a review of minimization methods, see~\cite{SZSVKATPAMI:2008}.

We minimize the energy in Fig.~\ref{eq:main-energy} with the jump move algorithm~\cite{Veksler:phd}, but see~\cite{SZSVKATPAMI:2008} for a review of minimization methods.  
Given a labeling  $\bx$, we say that  $\bx'$ is a jump move from $\bx$ if whenever $x_p \neq x'_p$, then $x'_p = x_p+1$, i.e. a jump move allows any pixel to increase its label by 1.  There is an exponential number of jumps. An optimal jump move decreases the energy the most. It is computable with a graph-cut~\cite{Kolmogorov:PAMI04} if submodularity holds~\cite{Boros01pseudo-booleanoptimization}. However, the smoothness terms $s$ are nonsubmodular  when $x_p \neq x_q$.
We tried QPBO~\cite{kolmogorov2007minimizing} for nonsubmoduar energies, but it did not work well. Instead,
 we replace nonsubmodular  terms by a submodular upper bound, enabling graph-cut minimization. An optimal move is not guaranteed, but the energy is guaranteed to decrease, see Sec.~\ref{sec:binary_energy_for_jump}.
 
 The jump move algorithm works as follows. We initialize all pixels to $0$. Then we perform jump moves until the labeling stops changing. We say that $(p,q)\in\calO$ is {\emph {activated}
if $x_p>x_q$ so that $o(x_p,x_q)$ contributes $-1$ to the energy. The more terms in $\calO$ are activated, the lower is the energy.
Consider Fig.~\ref{fig:jumps}.
 The first jump, Fig.~\ref{fig:jumps}c,  switches to label $1$ a segment of  pixels
 which have a low background probability and are also  aligned to the oriented occlusion boundaries, creating one instance. With only one instance, many $(p,q)\in\calO$ stay not activated. The second jump, Fig.~\ref{fig:jumps}d, assigns pixels in the front-most car to label 2, activating more $(p,q)\in\calO$ and decreasing the energy. The third jump, Fig.~\ref{fig:jumps}e,  increases by 1 the label of the frontal car, and the car behind it, and now most $(p,q)\in\calO$ are activated. Further jumps result in no changes.
 %If we try to segment another instance, it would align mostly to $(p,q)\not\in\calO$, and we would have to pay smoothness costs $s(x_p,x_q)$ along most of the segment boundary, increasing the energy. 

%TODO: If there is space, discuss the balance between smoothness and occlusion terms
 % In Eq.~\cref{eq:main-energy}, we we set $\lambda_s=10$ and $\lambda_o=30.$ This makes the absolute value of 

\subsection{Binary Energy for Jump Move and its Submodular Upper Bound}
\label{sec:binary_energy_for_jump}

We now explain the binary energy for the jump move, why it is not submodular~\cite{Boros01pseudo-booleanoptimization}, and how we construct its submodular upper bound.

Given a labeling $\bx$, a jump move either does not change the label of a pixel, or increases the label by 1. We  pose the problem of finding the optimal jump 
move as binary energy minimization. Let $\bx^c$ be the current labeling for which we wish
to find the optimal jump move. We introduce a binary variable $y_p$
for each pixel $p$, and collect  these variables into a vector $\by$. We define one-to-one correspondence between the set of jump moves from $\bx^c$ and the set of all
possible labelings of $\by$ as follows. For each labeling $\by$, let us denote the corresponding
jump move as $m(\bx^c; \by)$, defined as 

\begin{equation}
m(\bx^c; \by) =
    \begin{cases}
      x^c_p& \text{if  $y_p=0$}\\
      x^c_p+1 & \text{otherwise}
    \end{cases}       
\end{equation}
In words, label 0 is identified with  a pixel keeping its label, and label 1 with a pixel increasing its label by 1.

In the binary energy, let  $\hatu,\hatv,\hato$  denote the unary, pairwise smoothness and occlusion terms, which are derived from the corresponding terms of the multi-label energy in Eq.~(\ref{eq:main-energy}).

We define the unary terms of binary energy as
\begin{equation}
\hatu(0)=u(x^c_p), \; \hatu(1)=u(x^c_p+1).
\end{equation}
Given a pixel pair $(p,q)\in\calN$, we define the smoothness 
 pairwise terms of the binary energy as
\begin{align}
\hatv(0,0)  =&v(x^c_p,x^c_q)\\ \nonumber
\hatv(0,1)=&v(x^c_p,x^c_q+1)\\  \nonumber
\hatv(1,0)=&v(x^c_p+1,x^c_q)
\\\nonumber
\hatv(1,1)=&v(x^c_p+1,x^c_q+1).
\end{align}

Given a pixel pair $(p,q)\in\calO$, we define the occlusion
 pairwise terms of the binary energy as
\begin{align}
\hato(0,0)=&o(x^c_p,x^c_q)\\ \nonumber
\hato(0,1)=&o(x^c_p,x^c_q+1)\\ \nonumber
\hato(1,0)=&o(x^c_p+1,x^c_q)\\ \nonumber
\hato(1,1)=&o(x^c_p+1,x^c_q+1).
\end{align}

We define the binary energy for $\by$ as
%\vspace*{-1ex}
  \begin{equation} 
E_b(\by)=  \sum_{p \in \calP}  \hatu_p(y_p) + \lambda_v \sum_{(p,q)\in \calN} \hatv(y_p,y_q) + \lambda_o \sum_{(p,q)\in \calO}  \hato(y_p, y_q).
\label{eq:jump-energy}
\end{equation}

%Olga

 It is straightforward to check that $E_b(\by) = E({\bf{m}}(\bx^c,\by))$. Therefore the optimal jump
 move from $\bx^c$ corresponds to the labeling $\by^*$  minimizing Eq.~(\ref{eq:jump-energy}).

 If a binary energy is submodular~\cite{Boros01pseudo-booleanoptimization} then it can be optimized with a graph-cut~\cite{Kolmogorov:PAMI04}. For submodularity to hold, there is no conditions on the unary terms, but  each pairwise term $f$ must satisfy
\begin{equation}
    f(0,0)+f(1,1) \leq f(0,1) + f(1,0).
\end{equation}

It is straightforward but tedious to check that the occlusion pairwise terms $\hato$ are submodular in all cases. Unfortunately, the smoothness term $\hatv$ is not submodular for any pair of pixels $(p,q)\in\calN$  such that $x^c_p + 1= x^c_q.$  Indeed, in this case
\begin{eqnarray}
    \hatv(0,0)+\hatv(1,1)=&v(x^c_p,x^c_q) +v(x^c_p+1,x^c_q+1) \\ \nonumber
    = &  1+1   \\ \nonumber
    >&v(x^c_p,x^c_q+1) +v(x^c_p+1,x^c_q)  \\ \nonumber
    = &1+0\\ \nonumber
   =& \hatv(0,1)+\hatv(1,0) \nonumber
\end{eqnarray}

To make optimization feasible with a graph-cut, we replace the energy in Eq.~(\ref{eq:jump-energy}) with its submodular upper bound. For any $(p,q)\in\calN$, if $x^c_p+1=x^c_q,$ we replace $\hatv$ corresponding to this pixel pair by a constant term $c$ which always takes value 1 and, therefore, is submodular. 
%That is $c(a,b)=1$ for any values of $a,b$. 
%In practice, it is equivalent (the energies are equal up to a constant) to omitting the smoothness term $\hatv$ for any pixels $(p,q)\in\calN$ s.t. $x^c_p+1=x^c_q.$

Intuitively, our approximation means the following. If in the current labeling $\bx^c,$ we have a neighboring pair of pixels whose label differs by exactly 1, then the jump move is not able to `see' that it can achieve  a lower-energy labeling by increasing the smaller label by 1. Minimizing an upper bound as opposed to the exact binary energy means that the optimal jump move is not guaranteed if there are neighboring pixel pairs in the current labeling whose labels  differ by exactly 1. 
However  the energy is guaranteed to decrease (or stay the same).

For details on how to construct the graph for submodular binary function minimization, see~\cite{Kolmogorov:PAMI04}. We use the graph-cut/max-flow algorithm~\cite{boykov2004experimental} to compute the minimum cut.

\section{Joint OOSIS Metric: OAIR Curve}
\label{sec:OOSIS_metric}

\begin{figure}
  \centering
  \includegraphics[width=0.4\textwidth]{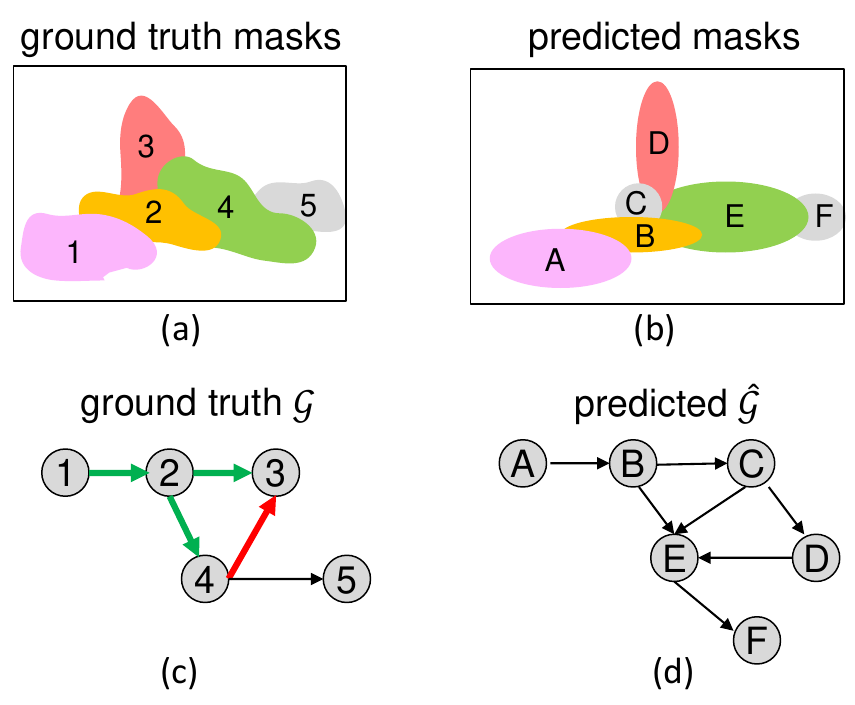}
  \caption{Occlusion graphs and matched regions for joint OOSIS metric computation: (a) and (b) show ground truth and predicted masks. Matched masks between the ground truth and the prediction are shown with the same color, unmatched masks ($5, C, F$) are in gray; (c) and (d) are the ground truth and the predicted occlusion graphs.
  }
   \label{fig:occlusiongraph}
\end{figure}

Our new joint OOSIS metric is based on Accuracy vs. Recall curve. We call it OAIR curve because \emph{{O}}cclusion order influences {\emph {A}}ccuracy, and \emph{I}nstance masks influence {\emph {R}}ecall.
We construct occlusion graphs $\mathcal{G}$, based on ground truth, and $\hat{\mathcal{G}}$, based on the predicted instances and occlusion relations.
In an occlusion graph, instances are nodes and directed edges connect occluders to their 
occludees. 
Fig.~\ref{fig:occlusiongraph} is a toy example for  $\mathcal{G}$ and  $\hat{\mathcal{G}}$.  In the ground truth graph, there are 5 instances, and in the predicted graph, there are 6 instances. 

To establish the correspondence between predicted and ground-truth instances, we follow a standard instance segmentation procedure~\cite{DBLP:journals/corr/LinMBHPRDZ14,Qi_2019_CVPR}. We begin by selecting the predicted instance with the highest confidence and seek a matching ground-truth instance with the same semantic class and the highest Intersection-over-Union (IoU). If the IoU exceeds a minimum threshold, the instances are matched, and we proceed to the next predicted instance. By raising the IoU threshold, we can reduce the number of matched instances. We also have a minimum confidence score below which predicted instances are not considered. 
In Fig.~\ref{fig:occlusiongraph}, matched regions between the ground truth and the predicted results are shown with the same color.

After matching, we consider two nodes $g_1$ and $g_2$ in $\mathcal{G}$, where $g_1$ occludes $g_2$, i.e. endpoints of edges. If both $g_1$ and $g_2$ have a matched detection in $\hat{\mathcal{G}}$, we label the pair as ``recovered'', otherwise as ``missed''. 
In  Fig.~\ref{fig:occlusiongraph}(c), the recovered pairs are indicated by thick arrows, and the missed pairs by thin arrows. There are 5 edges in the ground truth graph, but only 4 are recovered. 

We define a pair as ``correctly ordered'' if there is a directed path from the matched detected instance of $g_1$ to that of $g_2$ in $\hat{\mathcal{G}}$ and no path vice versa, i.e. one single order in the correct direction exists between the two. 
 In Fig.~\ref{fig:occlusiongraph}(c), the correctly ordered recovered edges are in green, and the incorrectly ordered ones are in red. For the recovered edges $(1,2)$ and $(2,4)$ the paths validating their correct order are of length 1. For the recovered edge $(2,3)$ the path validating the correct order is of length 2, i.e.   $B\rightarrow  C\rightarrow D$. 

We compute the  recall and accuracy, where recall is the ratio of recovered pairs to the total number of pairs, and accuracy is the ratio of correctly ordered pairs to the total number of recovered pairs.
In Fig.~\ref{fig:occlusiongraph}, the recall is $4/5$, and the accuracy is $3/4$. The pair $(4/5,3/4)$ is one point for the OAIR curve. 
We plot the OAIR curve by varying thresholds in  the matching criteria, either based on IoU or confidence score. 
Our OAIR curve is similar to what~\cite{wang2016doc} use for  occlusion boundaries.

\section{Experiments}
\label{sec:experiments}
In the rest, we name our method as \textit{Joint-Labeling}.

\subsection{Direct Training}
\label{sec:naive_approach}

\begin{figure}
  \centering
\includegraphics[width=0.5\textwidth]{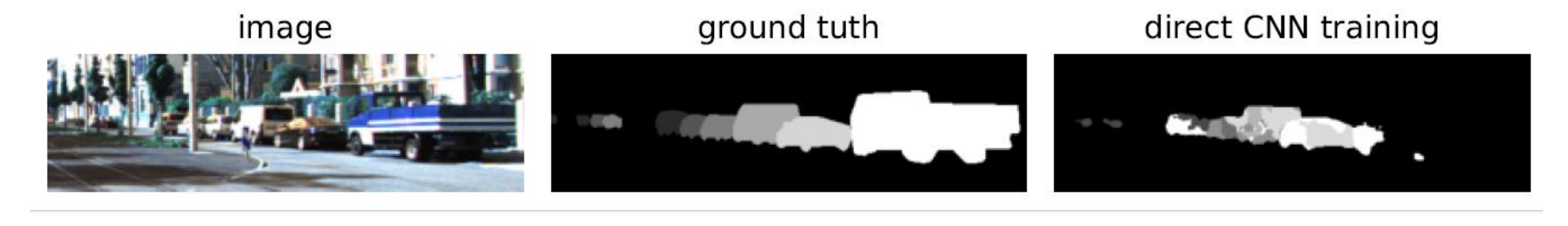}
  \caption{Relative depth maps  for direct training. 
  }
   \label{fig:easy-baselines}
\end{figure}

Since OOSIS has not been approached post-deep learning, first we check if directly training CNN on occlusion order labels works.
We create the ground truth as follows. All background pixels are labeled with $0$.
All pixels in an instance are labeled with $i$ if the maximum label of its occludees is $i-1$. Then we
 train a pixel-level CNN to recover these labels using cross-entropy. The results of direct training are poor, see Fig.~\ref{fig:easy-baselines}. This is not surprising, since OOSIS is a harder task than standard instance segmentation, for which there is no direct training approach. However, direct training for OOSIS is worth trying since the ground truth for OOSIS has extra (occlusion) information and occlusion-based labels are meaningful. 
The performance of direct training is so poor that we do not consider it as a baseline and do not include it in quantitative evaluation.

\subsection{Baselines} 
\label{sec:experiments:baselines}
In the absence of deep learning methods for OOSIS, we construct strong baselines by integrating methods addressing its distinct facets: instance segmentation and occlusion ordering of instances. 
To derive instances, we
 use either instance or panoptic segmentation. Subsequently, we order the instances by either applying pairwise CNN classifiers, trained specifically on each dataset, or depth from a monocular estimator averaged on instance pixels.

For the first component, we use state-of-the-art models PanopticMask2Former~\cite{cheng2021mask2former}, the contour-based E2EC~\cite{zhang2022e2ec}, and the classic Mask-RCNN~\cite{wu2019detectron2}. This selection encompasses diverse approaches including panoptic vs. instance segmentation and mask vs. contour-based segmentation. 

To order extracted instances, we use InstaOrderNet~\cite{lee2022instance} and OrderNet~\cite{zhu2017semantic}. These models are trained on either the KINS or COCOA ground truth to establish occlusion order for a pair of input instances. For alternative ordering based on the average monocular depth, 
for  COCOA 
we use MiDaS~\cite{Ranftl2022} depth estimator, since COCOA lacks depth data, and MiDaS was trained on a combination of diverse datasets. For the KINS, we use HR-Depth~\cite{lyu2021hr}, trained on KITTI~\cite{Geiger2013IJRR}, the parent dataset of KINS, yielding a stronger performance than MiDaS.
%a baseline that utilizes existing techniques. Specifically, it involves using an instance segmentation method to generate instance predictions and then applying a pairwise occlusion ordering CNN directly on neighboring instances to obtain the occlusion order without any further processing or mask overlap removal, as is the case with our top-down approach.

% \begin{wrapfigure}{r}
% {0.7\textwidth}
%   \begin{center}
%     \includegraphics[width=6cm]{Fig/qual_cvpr.pdf}
%     \end{center}
%     \caption{Relative depth map visualization of OOSIS for the best baseline and our approach. For the best baseline, note incorrect relative ordering for smaller objects, unlike our results.}
%     \label{fig:qual}    
% \end{wrapfigure}

\subsection{Experimental Settings}
\label{sec:experiments_settings}
\textbf{Datasets:} We use KINS dataset~\cite{Qi_2019_CVPR}, which has 14,991  driving scenes, and COCOA dataset~\cite{zhu2017semantic}, which has 5,073 natural scenes. They provide modal and amodal instance masks with occlusion order. We do not use amodal masks.

\textbf{Implementation Details}:  
%Olgaremoved: All models are only trained using modal masks and occlusion orders.
For E2EC~\cite{zhang2022e2ec}, we use the official implementation and configuration with DLA-34~\cite{yu2018deep} backbone. For Mask-RCNN, we use implementation in~\cite{wu2019detectron2} with configurations in~\cite{Qi_2019_CVPR} for KINS and the default configurations for COCOA~\cite{lin2014microsoft}, with  Resnet50~\cite{he2016deep} backbone. For PanopticMask2Former~\cite{cheng2021mask2former} we use the implementation in~\cite{wu2019detectron2}.
%We adjust the number of training epochs according to the dataset size. 
The panoptic model is trained with all non-instance  image parts grouped into a mother stuff class (background), since the datasets do not contain annotations for stuff classes, and our method and the baselines do not use such data in training. Our boundary and semantic segmentation network is a modified PSPNet~\cite{zhao2017pspnet}, with a ResNet50 backbone, trained for 200 epochs with   225x225 crops. For a fair comparison, all methods are evaluated on 2048x615 resolution for KINS and the original resolution for COCOA. For P2ORM~\cite{qiu2020pixel}, we use the official implementation. We take the pre-trained models of HR-Depth, MiDaS, InstaOrderNet, and OrderNet for each dataset from the official repositories. 

\begin{figure}
  \centering
  \includegraphics[width=0.5\textwidth]{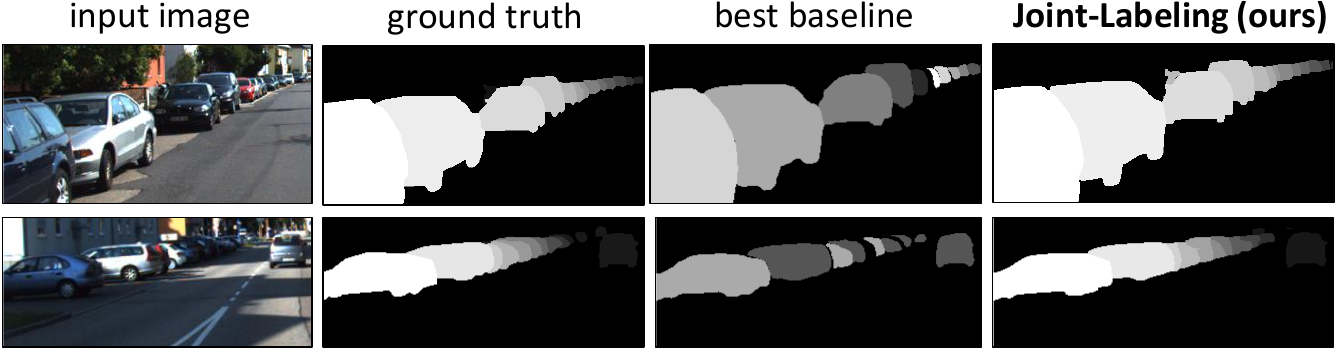}
  \caption{Relative depth maps of OOSIS for the best baseline and our approach. For the best baseline, note incorrect relative ordering for smaller objects, unlike our results.}
   \label{fig:qual}
\end{figure}

\subsection{Mask Evaluation Metrics}
\label{sec:metrics}
We use two  metrics, AP and WC,  for mask evaluation.

{\bf{Average Precision Metric (AP)}:}
Average precision (AP)~\cite{Everingham10}, inspired by object detection, is the most commonly used metric for semantic instances.
 There are multiple works that analyze shortcomings of AP
~\cite{dave2021evaluating,oksuz2018localization,zhang2015monocular,bai2017deep,jena2022beyond}.
One problem with AP is that it does not penalize
overlapping instances or a large number of instance predictions, provided that the instances   are ranked in the correct order~\cite{jena2022beyond}. 
 Thus the unambiguous approaches, i.e. approaches which predict a single instance label per pixel,  are at a disadvantage.  
Our approach is unambiguous. 

Detection-based instance segmentation methods   give a confidence score 
to predicted instances. In contrast, other instance segmentation methods, such as bottom-up~\cite{bai2017deep}, do not produce confidence scores.  
Therefore another serious drawback of AP metric~\cite{bai2017deep} is that it depends on the confidence score assigned to instances. If we take the same instance segmentation and assign different confidence scores to the instances, AP scores will be different, even though the segmentation is exactly the same. 
This is an unappealing property for unambiguous instance segmentation, where the confidence scores are not required, as all the instances appear  in the final result. 

Since our approach does not produce confidence scores, we have to come up with ad-hock confidence scores to employ AP metric. 
In~\cite{bai2017deep} they use simple heuristics when a metric requires instance confidences. We follow this practice. Specifically, we assign a score to each instance by summing the predicted occlusion boundary probabilities on the border and inside of the instance, and subtracting the second from the first.
%for its border and interior pixels and computing their difference. 
This score rewards instances if their borders match the predicted boundaries 
and they do not have high-probability boundaries inside, which indicates that they should not be further divided.
It also gives higher scores to larger instances, and these tend to be more reliable.

{\bf{Weighted Coverage Score (WC)}}
In~\cite{jena2022beyond} they propose metrics which are more appropriate for the approaches producing overlapping instances. However, since our approach does not produce overlapping instances, we advocate the use of the Weighted Coverage metric~\cite{pmlr-v22-tarlow12a,silberman2014instance}. 

Let $G=\{r_1,...,r_k\}$ be the set of ground truth instances, and $S=\{s_1,...,s_l\}$ be the set of segmented instances in an image. 
%Given a pair of regions $a,b$,  the intersection over union score is defined as
% \begin{equation}
% IoU(a,b) = \frac{a \cap b}{a\cup b}    
% \end{equation}
The Weighted Coverage (WC) metric is defined as 
\begin{equation}
   WC(G,S) = \frac{1}{n} \sum_{i=1}^{k} |r_i| \max_{j=1,...,l} IoU(r_i,s_j),
\end{equation}
where $n$ is the number of pixels in the image, $|r_i|$ is the size of region  $r_i$, and $IoU$ is the intersection over union score.
Intuitively, WC matches each ground truth instance with a segmented instance of the largest overlap, and adds to the score the size of the ground truth region, weighted by the goodness of this overlap (where goodness is equal to the IoU score).  The best value of WC is 1, obtained when the set of segmented instances contains exactly the same segments as the ground truth. WC metric does not depend on the confidence of a segmented instance, which is an intuitive property for evaluating unambiguous methods.

Note that using WC metric makes sense only for unambiguous instance segmentation methods, i.e. each pixel is assigned to at most one instance. Otherwise, one can take  $S$ to be equal to the set of all possible segments, automatically achieving the best WC score of 1.

\subsection{Evaluation}
\label{sec:experiments_evaluation}
{\bf{OAIR Curves Evaluation:}}
\begin{figure*}
  \centering
  \begin{subfigure}{0.88\textwidth}
      \includegraphics[width=\textwidth]{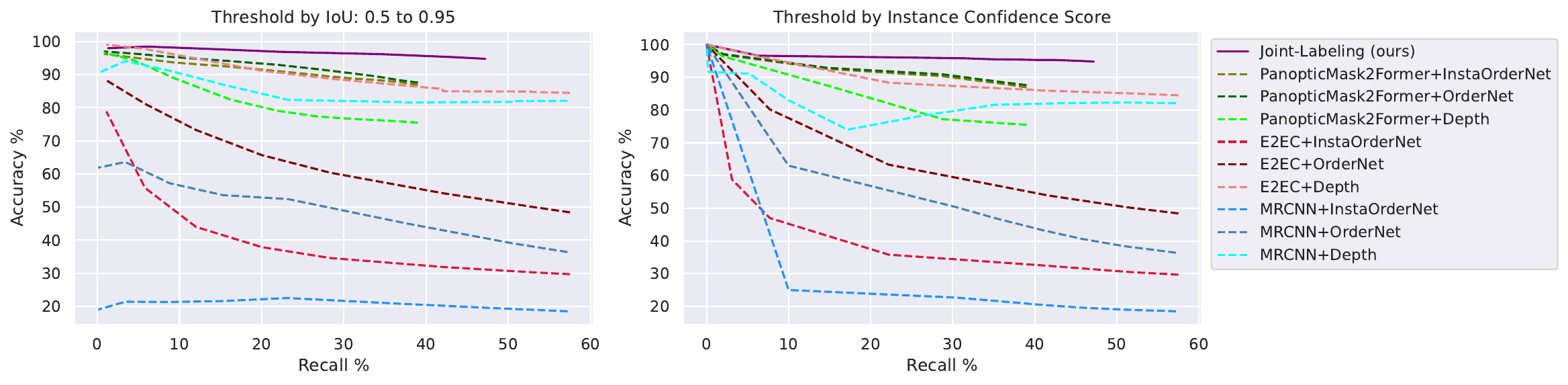}
      \caption{Our method compared to baselines on KINS.}
      \label{fig:oosis-vs-baseline-kins}
   \end{subfigure}
   \begin{subfigure}{0.88\textwidth}
      \includegraphics[width=\textwidth]{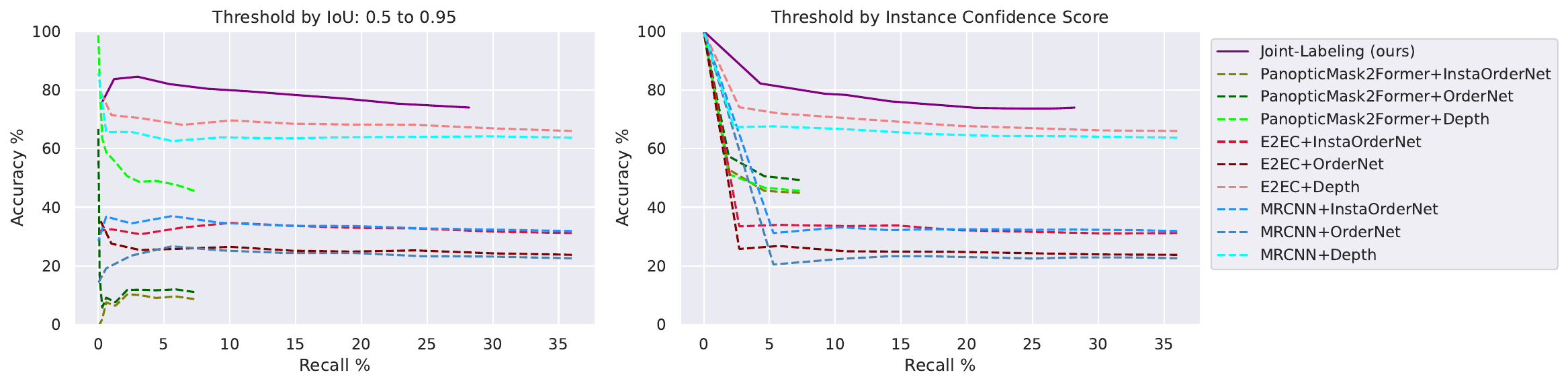}
      \caption{Our method compared to baselines on COCOA.}
      \label{fig:oosis-vs-baseline-cocoa}
   \end{subfigure}
  \caption{Left: OAIR curves for IoU thresholds ranging from 0.5 to 0.95 in steps of 0.05. Right: OAIR curves for varying minimum confidence score, with six thresholds covering 0 to the maximum recall of each method. A higher curve is better performance.}
\end{figure*}
Fig.~\ref{fig:oosis-vs-baseline-kins} gives OAIR curves for our method and the baselines on KINS. Our method has the highest accuracy for similar recalls, outperforming the best baseline by a margin of $\sim10\%$ in accuracy for similar recall values above $40\%$. Among the baselines,  PanopticMask2Former+OrderNet shows higher accuracy for recall values below $30\%$. However, it cannot achieve any higher recalls. Notably, while InstaOrder outperforms OrderNet on the pairwise occlusion ordering task~\cite{lee2022instance}, it performs slightly weaker than OrderNet when used with PanopticMask2Former for  OOSIS.
On the other hand, baselines based on E2EC and M-RCNN show higher recall but lower accuracy. Their higher recall is because these methods have overlapping predictions for instances instead of an unambiguous segmentation, where each pixel is assigned to at most one instance. Unlike them, the output of our method and PanopticMask2Former is unambiguous. Fig.~\ref{fig:oosis-vs-baseline-cocoa}  compares our method with the baselines using OAIR curves on COCOA. Our method performs significantly better than the baselines, with better accuracy across all recall values. Unlike for KINS dataset, baselines based on E2EC and M-RCNN  perform better than ones employing PanopticMask2Former on COCOA, mostly due to extracting higher-quality instances. 

{\bf{Ordering based on Monocular Depth vs. Occlusion:}}
Fig.~\ref{fig:oosis-vs-baseline-kins} shows the efficacy of occlusion for relative depth compared to relying on monocular depth.  KINS consists of driving scenes where some objects are at a considerable distance from the camera, and monocular depth is less reliable in establishing relative depth for distant objects. Notably, the plot shows not only the superior performance of our occlusion-based method over depth-based baselines but also the better performance of ordering through CNN classifiers, InstaOrderNet and OrderNet, compared to the depth baselines. On  COCOA (Fig. \ref{fig:oosis-vs-baseline-cocoa}), while depth-based baselines show better performance over the CNN classifiers, possibly due to dominance of closer objects in the indoor scenes, our method still outperforms depth-based ordering. This underscores the reliability of occlusion  compared to monocular depth for the specific task of depth ordering.

{\bf{Consistency of Occlusion Ordering}:}
\begin{table}
        \centering
        \caption{\% of 
        predicted instances involved in occlusion cycles.}
        \label{tab:cycles}
        %\begin{small}
        \begin{scriptsize}
        \begin{tabular}{lcc}
            \toprule
            \multirow{2}{*}{Model} & \multicolumn{2}{c}{\% of Instances in Cycles $\downarrow$}\\
            & KINS & COCOA\\
            \midrule
            PanopticMask2Former+InstaOrder & $0.2$ & $1.1$\\
            PanopticMask2Former+OrderNet & $< 0.01$ & $1.3$\\
            PanopticMask2Former+Depth & $0$ & $0$\\
            \midrule
            E2EC+InstaOrder & $32.0$ & $62.5$\\
            E2EC+OrderNet & $30.3$ & $75.9$\\
            E2EC+Depth & $0$ & $0$\\
            \midrule
            M-RCNN+InstaOrder & $42.1$ & $66.6$\\
            M-RCNN+OrderNet & $42.3$ & $79.9$\\
            M-RCNN+Depth & $0$ & $0$\\
            \midrule
            \midrule
            Joint-Labeling (Ours) & $0$ & $0$\\
            \bottomrule
            \end{tabular}
            %\end{small}
            \end{scriptsize}
\end{table}

Let us call an OOSIS method globally consistent if it produces a cycle-free occlusion graph. Although cycles in a ground truth occlusion graph are possible, they are rare. The ground truth for both KINS and COCOA is  cycle-free. Thus predicted cycles on KINS and COCOA are errors, and not only undesirable, but also make it impossible to derive the relative depth order for any pair of instances in a cycle. Fig.~ \ref{tab:cycles} evaluates the baselines and our approach by reporting the percentage of predicted instances involved in occlusion cycles.
The baselines based on pairwise classifiers, InstaOrderNet and OrderNet, output cycles. Those based on E2EC and M-RCNN perform especially poorly. Those based on PanopticMask2Former show fewer cycles probably due to unambiguous instance masks. Our approach and depth-based baselines are cycle-free by design, giving a globally consistent ordering. Thus not only our approach performs better in terms of OAIR curves but its cycle-free output guarantees a globally consistent ordering.

{\bf{Semantic Instance Mask Quality}:}
\begin{table}
\setlength{\tabcolsep}{1pt}
  \caption{Instance segmentation quality by PanopticMask2Former vs our approach.}
  \label{tab:instance-seg-ours}
  \centering
  %\begin{small}
  \begin{scriptsize}
  \begin{tabular}
  {ccccc}
    \toprule
    \multirow{3}{*}{Model} & \multicolumn{4}{c}{Unambiguous Instance Segmentation} \\
    & \multicolumn{2}{c}{KINS} & \multicolumn{2}{c}{COCOA} \\
    \cmidrule(r){2-5}
    & $mAP_{0.5:0.95} \uparrow$ & WCS $\uparrow$ & $mAP_{0.5:0.95} \uparrow$ & WCS $\uparrow$\\
    \midrule
     PanopticMask2Former (same score) & 15.4 & 74.6 & 3.7 & 47.3\\
     PanopticMask2Former (ad-hoc score) & \textbf{21.6} & 74.6 & 7.4 & 47.3\\
    \midrule
    Joint-Labeling (Ours) & \textbf{21.6} & \textbf{77.4} & \textbf{9.1} & \textbf{61.2}\\
    \bottomrule
  \end{tabular}
%\end{small}
\end{scriptsize}  
\end{table}
We now evaluate the quality of instance masks using standard metrics. Occlusion ordering is not considered for this evaluation.
An instance segmentation method is unambiguous if each pixel is assigned to at most one instance. Our approach is unambiguous, whereas most instance segmentation methods are ambiguous~\cite{jena2022beyond}. 
As~\cite{bai2017deep,jena2022beyond} point out, standard metrics favor ambiguous methods.
Thus, for a fair comparison, we compare only unambiguous methods, i.e. PanopticMask2Former baselines. 
We report metrics: $mAP_{0.5:0.95}$~\cite{DBLP:journals/corr/LinMBHPRDZ14}~\cite{Qi_2019_CVPR} and Weighted Coverage Score (WCS)~\cite{silberman2014instance}. The latter does not need instance confidence scores while the former does.
%The former is  widely-used  for instance segmentation~\cite{DBLP:journals/corr/LinMBHPRDZ14}~\cite{Qi_2019_CVPR}. The latter is more equitable when comparing bottom-up approaches vs. top-down ones, as it removes the need for predicting a mask confidence score~\cite{bai2017deep}. However, it is only applicable when the predicted masks are unambiguous~\cite{bai2017deep}.
%~\cite{bai2017deep, silberman2014instance}.

Tab.~\ref{tab:instance-seg-ours} shows the results. For PanopticMask2Former we report two versions different only in the confidence score computation. The first version assigns the same score to all masks. The second version computes an ad-hoc confidence score similar to~\cite{cheng2021mask2former} by multiplying the mask confidence score and class confidence score of each pixel.
%predicted by internal parts of the architecture. %TODO 
These methods only differ in the mAP score. 
Our method is superior to both versions of PanopticMask2Former on both KINS and COCOA datasets, and on both metrics, highlighting the accuracy of our instance masks. Therefore, our OOSIS approach, when focusing just on instance masks, can be regarded as a new method for instance segmentation, albeit requiring occlusion information for training. Note that PanopticMask2Former is not trained with occlusion information, but it has a more powerful transformer based architecture, while our method is based on PSPNet. We chose PanopticMask2Former despite its more advanced architecture for mask quality comparison as PanopticMask2former is state-of-the-art in unambiguous instance segmentation. 
%Resolving the overlap by High Confidence yields superior results vs. other  policies.  E2EC is better than M-RCNN in all cases. Our bottom-up approach achieves the best Weighted Coverage Score (WCS), despite its lower mAP. This is due to the confidence score assignment influence, see~\cite{bai2017deep, jena2022beyond} and Supplementary.
%Olgaremoved: This is due to the confidence score assignment approach employed, as discussed in~\cite{bai2017deep}. 
%While an improved scoring may increase mAP for bottom-up approaches, WCS  is  a more equitable comparison.
%, where our bottom-up method exhibits better performance than our top-down approaches. 
%The results highlight the effectiveness of the High Confidence overlap resolution, and the strength of our bottom-up approach  when compared with the recent  powerful instance segmentation E2EC.

\begin{table}
\setlength{\tabcolsep}{1pt}
\centering
\begin{subtable}[b]{0.45\linewidth}
            \begin{small}
            \begin{tabular}{cccc}  
                \toprule
                Model & \multicolumn{3}{c}{Oriented Occ Boundary}\\ 
                & ODS$\uparrow$ & OIS$\uparrow$ & AP$\uparrow$\\
                \midrule
                P2ORM & 77.8 & 80.6 & 79.4\\
                Ours    & \textbf{84.5} & \textbf{86} & \textbf{88.3}\\
                \bottomrule    
              \end{tabular}
              \end{small}
\caption{The performance of our oriented occlusion boundary model vs. the state-of-the-art, P2ORM. OIS, ODS, are F-measures based on the best threshold per image, or for the whole dataset, respectively. AP is average precision. All are based on POR curves for NMS-applied oriented occlusion boundaries on KINS.}
\label{tab:occlusion-edge}
\end{subtable}
\hspace{3ex}
\begin{subtable}[b]{0.45\linewidth}
\centering
    \begin{small}
        \begin{tabular}{c|c}
            \toprule
            Loss & AP$\uparrow$\\
            \midrule
            Dice++~\cite{dicepp} & 91.7 \\
            Tversky++~\cite{dicepp} & 91.2\\
            ComboLoss~\cite{taghanaki2019combo} & 88.9\\
            \midrule
            WeightedCE & \bf{92.4}\\
            \bottomrule
        \end{tabular}
    \end{small}
    \caption{Performance of $b$ head of our boundary model using various loss functions in terms of AP metric on KINS.}
  \label{tab:ablation-boundary-loss}
\end{subtable}
\caption{}
\end{table}

%\subsection{Oriented Occlusion Edges of the Bottom-up Approach}
{\bf{Oriented Occlusion Boundary Model:}}
% \begin{table}[!t]
%             \centering
%             \caption{The performance of our oriented occlusion boundary model vs. the state-of-the-art, P2ORM. OIS, ODS, are F-measures based on the best threshold per image, or for the whole dataset, respectively. AP is average precision. All are based on POR curves for NMS-applied oriented occlusion boundaries on KINS.}
%             \label{tab:occlusion-edge}
%             \begin{small}
%             \begin{tabular}{cccc}  
%                 \toprule
%                 Model & \multicolumn{3}{c}{Oriented Occ Boundary}\\ 
%                 & ODS$\uparrow$ & OIS$\uparrow$ & AP$\uparrow$\\
%                 \midrule
%                 P2ORM & 77.8 & 80.6 & 79.4\\
%                 Ours    & \textbf{84.5} & \textbf{86} & \textbf{88.3}\\
%                 \bottomrule    
%               \end{tabular}
%               \end{small}
% \end{table}
Although CRF-based labeling (Sec.~\ref{sec:instance_occlusion_labeling}) is crucial for OOSIS, it heavily relies on the quality of the occlusion boundaries (Sec.~\ref{sec:instance_occlusion_labeling}). We now evaluate our occlusion boundary model against the state-of-the-art method, P2ORM~\cite{qiu2020pixel}, which outperforms other work~\cite{wang2016doc,lu2019occlusion,wang2019doobnet}. We use the standard metrics based on Precision vs. Recall curves (POR)~\cite{wang2019doobnet}. 
%~\cite{qiu2020pixel,wang2019doobnet}. 
Tab.~\ref{tab:occlusion-edge} shows our superior performance in all metrics by a margin, underscoring the effectiveness of our novel boundary model.  

%It is worth noting that P2ORM and previous works were originally developed for extracting all oriented occlusion edges in an image, while in our case, we only focus on the edges relevant to objects with classes of interest. As shown in Table \cref{tab:occlusion-edge}, P2ORM performs weaker when trained and tested on this specific task.
%Recall is the ratio of ground-truth occlusion boundaries recovered and precision is the ratio of correctly oriented occlusion boundaries w.r.t all detected boundaries. 

% Supplementary materials contain additional details of our method, experiments, ablation studies, and experimental results. 

\subsection{Ablation Studies}
\label{sec:experiments_ablation}
{\bf{Occlusion Boundary Loss}:}
% \begin{table}[]
%     \centering
%     \caption{Performance of $b$ head of our boundary model using various loss functions in terms of Average Precision metric on KINS.}
%     \begin{small}
%         \begin{tabular}{c|c}
%             \toprule
%             Loss & AP$\uparrow$\\
%             \midrule
%             Dice++~\cite{dicepp} & 91.7 \\
%             Tversky++~\cite{dicepp} & 91.2\\
%             ComboLoss~\cite{taghanaki2019combo} & 88.9\\
%             \midrule
%             WeightedCE & \bf{92.4}\\
%             \bottomrule
%         \end{tabular}
%     \end{small}
%     \label{tab:ablation-boundary-loss}
% \end{table}
Next, we show the effect of different losses for the $b$ head of our oriented occlusion boundary model instead of  Weighted Cross-Entropy. We compare with Dice++~\cite{dicepp} (with $\gamma=2$), Tversky++~\cite{dicepp} (with $\gamma=2$ and fine-tuned $\alpha=0.8, \beta=0.2$), and Combo~\cite{taghanaki2019combo} (with fine-tuned $\alpha=0.1$). We report the performance in terms of Average Precision in Tab.~\ref{tab:ablation-boundary-loss} on KINS.  Weighted CE performs the best. This ablation is interesting because  prior works advocate complex boundary losses such as Dice++, Tversky++, and Combo, while we show that simple cross-entropy with a carefully chosen weight parameter works better, at least for KINS dataset. 

{\bf{CRF Labeling}:}
%{\bf{Semantic Instance Mask Quality}:}
% \begin{table}
%   \caption{The quality of semantic instance segmentation by our different approaches.}
%   \label{tab:instance-seg-ours}
%   \centering
%   \begin{small}
%   \begin{tabular}{ccc}
%     \toprule
%     \multirow{3}{*}{Model} & \multicolumn{2}{c}{Unambiguous Instance Segmentation} \\
%     & \multicolumn{2}{c}{KINS}\\
%     \cmidrule(r){2-3}
%     & $mAP_{0.5:0.95} \uparrow$ & Weighted Coverage $\uparrow$\\
%     \midrule
%      PanopticMask2Former (Same Score) & 15.4 & 74.6\\
%      PanopticMask2Former (Ad-hoc Score) & 21.6 & 74.6\\
%     \midrule
%     Joint-IO-Labeling (from predicted boundaries) & 21.6 & 77.4\\
%     Joint-IO-Labeling (from GT boundaries) & \textbf{89.2} & \textbf{98.3}\\
%     \bottomrule
%   \end{tabular}
% \end{small}
% \end{table}
\begin{table}
  \caption{Unambiguous semantic instance segmentation by our method applied on predicted vs. ground truth occlusion boundaries and semantic segmentation on KINS.}
  \label{tab:ablation-gt-boundary}
  \centering
  \begin{scriptsize}
  \begin{tabular}{lcc}
    \toprule
    Model & $mAP_{0.5:0.95} \uparrow$ & WCS $\uparrow$\\
    \midrule
    Joint-Labeling (predicted bndry/segm) & 21.6 & 77.4\\
    Joint-Labeling (GT bndry/segm) & \textbf{89.2} & \textbf{98.3}\\
    \bottomrule
  \end{tabular}
\end{scriptsize}
\end{table}
To emphasize the effectiveness of the second stage of our approach,  CRF labeling, we conduct the following ablation. We substitute the outputs of the initial stage of our method with ground truth occlusion boundaries and semantic segmentation and apply CRF labeling on them. The results are in Tab.~\ref{tab:ablation-gt-boundary} for the mask metrics and  in Fig.~\ref{fig:supp-ablation-oair-curves-from-gt} for OAIR curves for KINS dataset.  The resulting OAIR curves show high degrees of accuracy. Notable, for Fig.~\ref{fig:supp-ablation-oair-curves-from-gt} (left), no matter how low we set the IoU threshold for instance matching the recall never drops below $~80\%$ while maintaining a high accuracy as well. 

These results  demonstrate that the second stage of our approach (CRF-based labeling) performs exceptionally well, and the bottleneck  is the first stage, computing oriented occlusion boundaries and semantic segmentation. 
%Thus our potential for future improvement is in addressing this bottleneck.

\begin{figure*}
    \centering
    \includegraphics[width=0.8\textwidth]{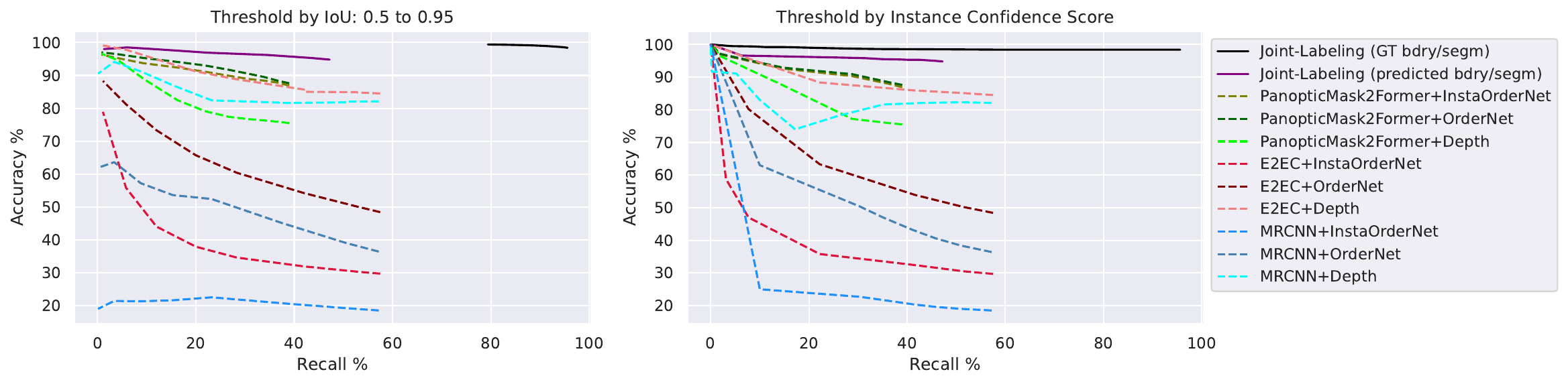}
    \caption{OAIR Curves for evaluation of our labeling approach and CRF-based optimization when applied on ground truth boundaries and semantic segmentation instead of predicted ones. }
    \label{fig:supp-ablation-oair-curves-from-gt}
\end{figure*}
\begin{figure*}
    \centering
    \includegraphics[width=0.9\textwidth]{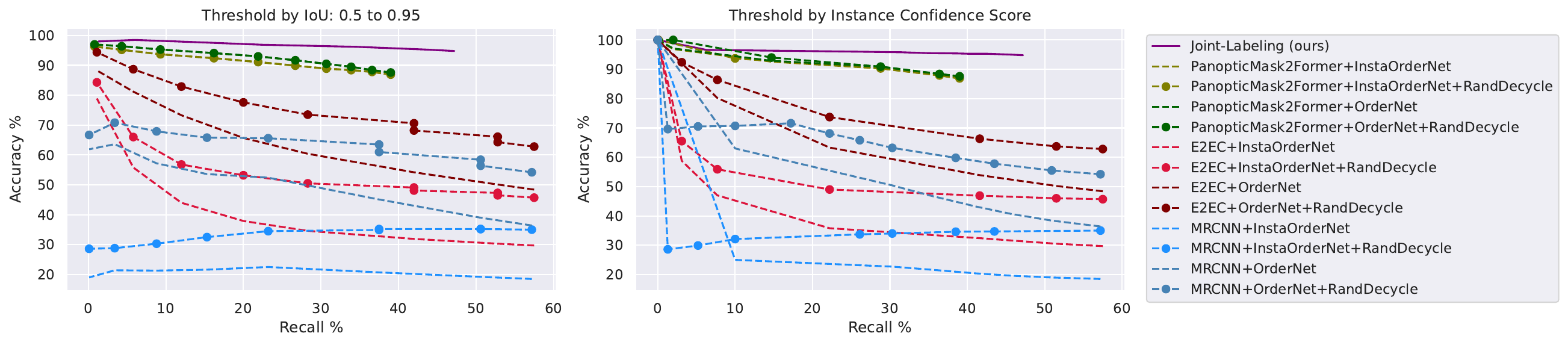}
    \caption{OAIR Curves for evaluation of baselines before and after post-hoc random cycle removal. Left: AOR curves for IoU thresholds ranging from 0.5 to 0.95 in steps of 0.05. Right: AOR curves for varying minimum confidence score, with six thresholds covering 0 to the maximum recall of each method. A higher curve is better performance.}
    \label{fig:supp-ablation-decycling}
\end{figure*}

{\bf{Oriented Occlusion Boundaries Architecture:}}
In Tab.~\ref{tab:ablation}
we show the effect of the joint upsampling for heads $e$ and $b$. Joint upsampling improves the performance on AP metric of POR curves for oriented occlusion boundaries, and also on AP metric for just detecting boundaries, i.e. the $b$ branch performance alone. This indicates the efficacy of sharing information in upsampling by the logits of $e$ and $b$.
\begin{table}[h]
    \centering
    \caption{Ablation on the effect of joint upsampling of heads $b$ and $e$. All are for NMS-applied oriented occlusion boundaries on KINS.}
    \label{tab:ablation}
    \begin{small}
    \begin{tabular}{ccccc}  
        \toprule
        Our Upsampling & \multicolumn{3}{c}{Oriented Occ Bndry} & Bndry Detection\\
        Approach& ODS$\uparrow$ & OIS$\uparrow$ & AP$\uparrow$ & AP$\uparrow$\\
        \midrule
        Separate $b$ \& $e$ & 84.5 & 85.9 & 87.1 & 90.2\\
        Joint $b$ \& $e$  & 84.5 & \textbf{86} & \textbf{88.3} & \textbf{92.4}\\
        \bottomrule    
    \end{tabular}
    \end{small}
\end{table}

{\bf{Random Decycling for Baselines:}}
The presence of cyclic occlusion orderings is highly undesirable. While our approach and depth-based baselines guarantee a cycle-free occlusion ordering, other baselines have numerous  cycles in  their occlusion order prediction, see Tab.~\ref{tab:cycles}. 
One idea is post-hoc decycling on the  occlusion order graphs produced by these baselines. However, it is hard to come up with an efficient intelligent strategy for cycle removal. For example, one may want to remove the smallest set of edges which results in an acyclic ordering graph.  However, this is an NP-hard 'feedback arc set' problem. Instead, we investigate decycling by applying a post-hoc random cycle removal as follows. We find a cycle, remove a random edge from it and check if there are still any cycles left. We repeat this procedure until we get an acyclic graph.

Fig.~\ref{fig:supp-ablation-decycling} is the performance results of this technique on the KINS dataset. Our approach remains the top performer in terms of accuracy at similar recall values. Baselines using PanopticMask2Former do not show a great improvement, as expected, due to their low number of cycles, see Tab.~\ref{tab:cycles}. Baselines using E2EC and M-RCNN enjoy a significant boost in their performance when such a decycling is applied. However, all baselines show weaker performance than our approach even after such post-processing.

{\bf{Effect of $\lambda$:}}
\begin{figure}
    \centering
    \includegraphics[width=0.5\textwidth]{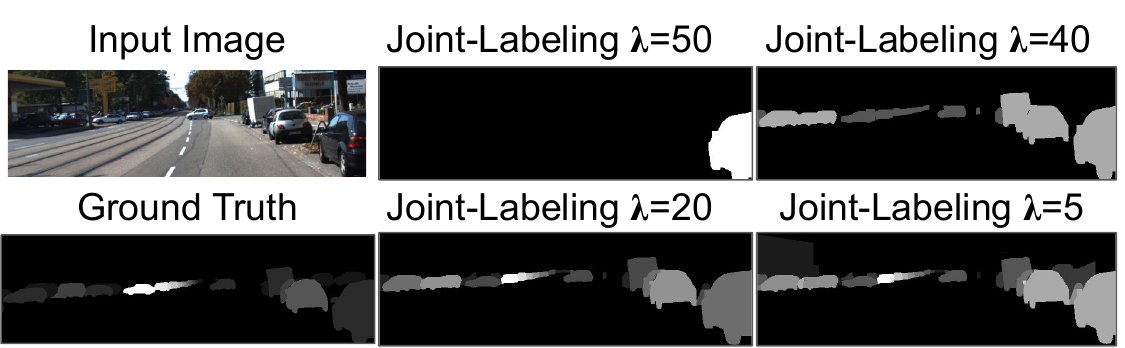}
    \caption{The effect of parameter $\lambda$ in our method. While large values cause the model to not detect instances ($\lambda=50$) or merge several instances into one instance ($\lambda=40$), a very low lambda can create noises in the output or cause the false detection of non-existing instances ($\lambda=5$). An appropriate $\lambda=20$ can work well and prevent these artifacts.}
    \label{fig:lambda-effect}
\end{figure}
In Fig.~\ref{fig:lambda-effect}
we investigate the impact of the hyperparameter $\lambda$ in the second stage of our approach, i.e. CRF-based labeling. Parameter $\lambda$  controls the weight of the smoothness term. Higher values of $\lambda$ tend to hinder instance detection, e.g. instances are not detected with $\lambda=50$, or multiple instances are merged, as observed for the left row of cars with $\lambda=40$. Conversely, excessively low values of $\lambda$ can introduce noise in the output or yield erroneous identification of non-existent instances, such as identifying the bus station as a vehicle with $\lambda=5$. 
Tab.~\ref{tab:instance-seg-param-ablation} shows quantitative evaluation  the $\lambda$  on the KINS dataset. 
Optimal performance is achieved at $\lambda=20$.

\begin{table}
  \caption{Instance segmentation quality by our approach using different $\lambda$ parameters.}
  \label{tab:instance-seg-param-ablation}
  \centering
  \begin{tabular}{ccc}
    \toprule
    \multirow{3}{*}{Model} & \multicolumn{2}{c}{Unambiguous Instance Segmentation} \\
    & \multicolumn{2}{c}{KINS} \\
    \cmidrule(r){2-3}
    & $mAP_{0.5:0.95} \uparrow$ & WCS $\uparrow$\\
    \midrule
    Joint-Labeling ($\lambda=50$) & 2.9 & 57.9 \\
    Joint-Labeling ($\lambda=40$) & 14.1 & 65.4 \\
    Joint-Labeling ($\lambda=20$) & \textbf{21.6} & \textbf{77.4} \\
    Joint-Labeling ($\lambda=10$) & 21.3 & 77.3 \\
    Joint-Labeling ($\lambda=5$) & 20.7 & 77.3 \\
    \bottomrule
  \end{tabular}
\end{table}

\subsection{Compute}
\label{sec:experiments:compute}
All deep models, E2EC, Mask-RCNN, IntaOrder, OrderNet, and ours were trained and tested on a single GPU of NVIDIA RTX3090 using PyTorch. CRF optimization for our approach was done on CPU using Python and Jupyter Notebook environment. 
Training our PSPNet-based deep model for joint semantic segmentation and occlusion boundary detection took $\sim 12$ hours for KINS and $\sim 4$ hours for COCOA. For our CRF optimization, testing on a single image of KINS on CPU with no parallelization took $\sim 12$ seconds. This was $\sim 13$ seconds for COCOA. Training E2EC for KINS took $\sim 15$ hours for KINS and $\sim 4$ hours for COCOA. Testing each of InstaOrder or OrderNet on any  configuration pair took $\sim 1.5$ hours. We did not have to train InstaOrder and OrderNet as we used pre-trained networks (on both KINS and COCOA datasets).

\section*{Conclusions}

To enable 3D analysis of a single scene, we propose to solve the joint task of Occlusion-Ordered Semantic Instance Segmentation (OOSIS). Occlusion is a simpler cue than monocular depth and can be estimated more reliably, albeit it results in relative, as opposed to absolute depth. 
%OOSIS has not been addressed in the deep learning framework before. 
We develop an approach which jointly infers instances and their occlusions by formulating OOSIS as a labeling problem.
As strong baselines for OOSIS, we combine state-of-the-art methods which separately perform instance segmentation and separately extract occlusion order. We show that our joint OOSIS approach outperforms these strong baselines, highlighting the advantages of the joint formulation.

Our main limitation is that occlusions provide only a partial order. The relative depth between two instances not connected by a monotone chain is unknown. Another limitation is that we cannot handle objects mutually occluding each other.


\bibliographystyle{IEEEtran}
\bibliography{tpami_oosis}

\begin{IEEEbiography}[{\includegraphics[width=1in,height=1.25in,clip,keepaspectratio]{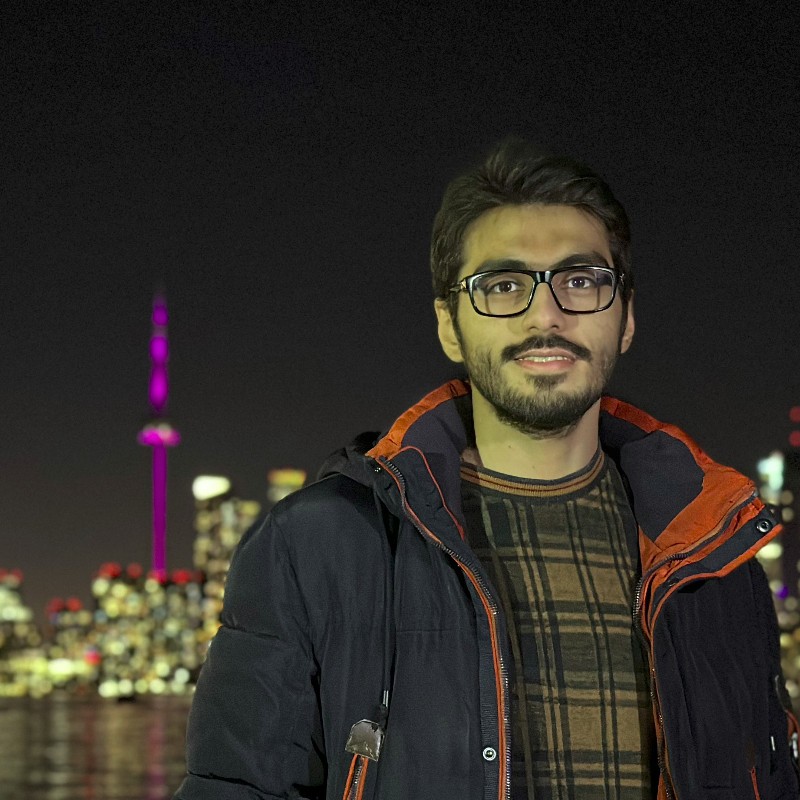}}]{Soroosh Baselizadeh} received his B.Sc. in Computer Engineering from Sharif University of Technology (SUT), Iran, and his Master's in Computer Science from the University of Waterloo, Canada. During his academic career, he collaborated with the Robust and Interpretable ML Lab at SUT and the Chair for Computer Aided Medical Procedures (CAMP) at the Technical University of Munich, Germany. His research spans various topics in machine learning and computer vision, with a focus on segmentation, explainable AI, and knowledge distillation. He is currently an ML Researcher and Engineer at Electronic Arts Inc., Canada, where he applies advanced machine learning techniques to enhance gaming experiences and develop innovative computer vision systems.
\end{IEEEbiography}

\begin{IEEEbiographynophoto}{Cheuk-To Yu} received his B. Math degree in computer science at Waterloo in 2022.  He is currently a MSc student in computer science at the University of Waterloo, focusing on computer vision and image segmentation. 
\end{IEEEbiographynophoto}


\begin{IEEEbiography}[{\includegraphics[width=1in,height=1.25in,clip,keepaspectratio]{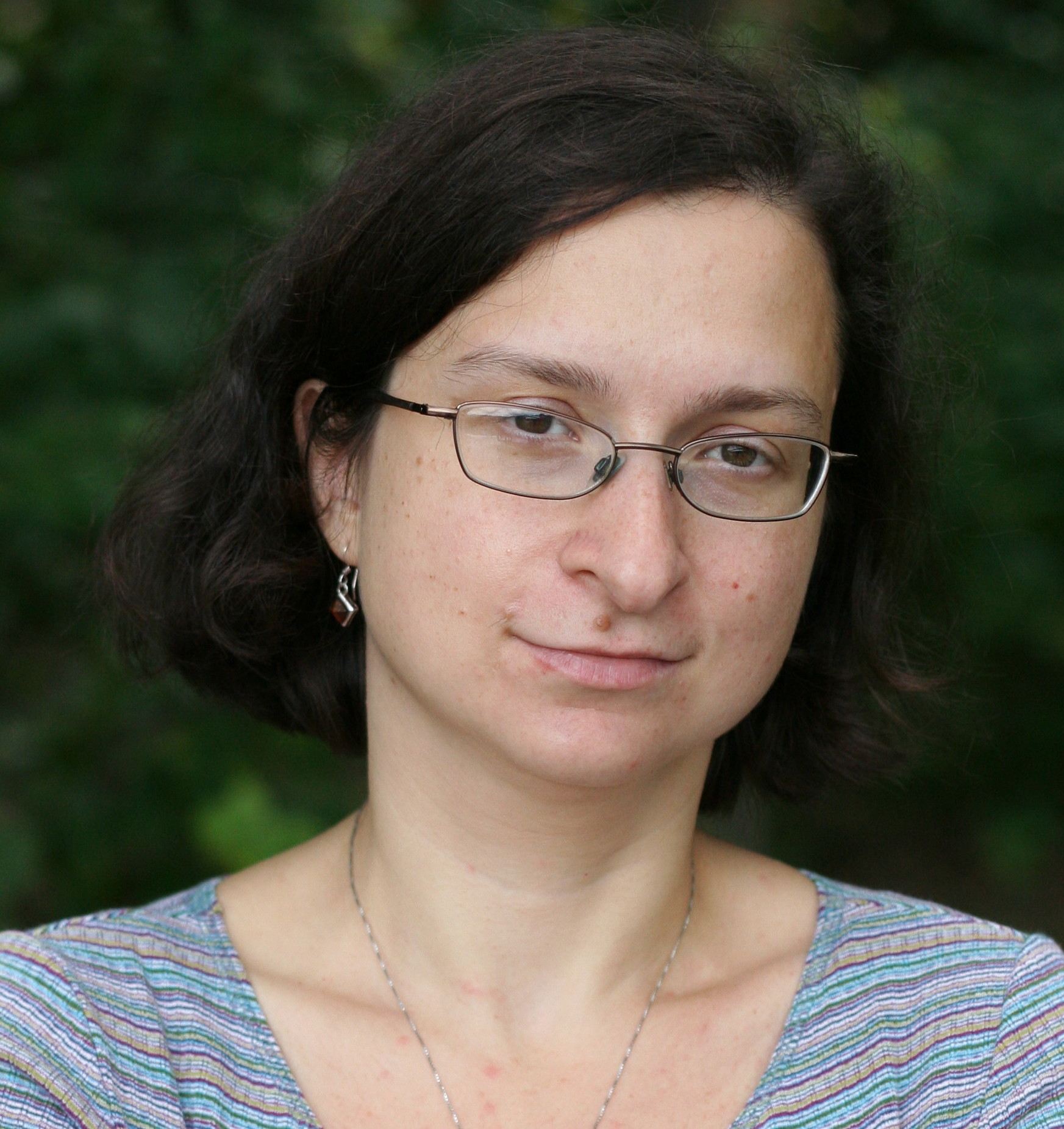}}]{Olga Veksler} received PhD degree from Computer Science Department, Cornell University in 1999. She is currently a full professor with the School of Computer Science, University of Waterloo.  She has served multiple times as an area chair for CVPR, ECCV, and ICCV. She is an associate editor for IJCV and TPAMI. She is a recipient of a "Test-of-Time Award" at the International Conference on Computer Vision, 2011.
\end{IEEEbiography}

\begin{IEEEbiography}[{\includegraphics[width=1in,height=1.25in,clip,keepaspectratio]{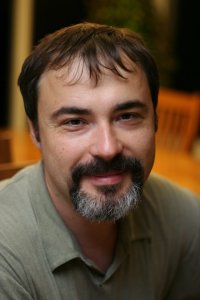}}]{Yuri Boykov}  is currently a professor at the Cheriton School of Computer Science, University of Waterloo, Canada. His
research interests include the area of computer vision and biomedical image analysis
with focus on modeling and optimization for
structured segmentation, restoration, registration, stereo, motion, model fitting, recognition,
photo-video editing and other data analysis
problems. He is an editor for the International
Journal of Computer Vision (IJCV). His work
was listed among the 10 most influential papers in the IEEE Transactions on Pattern Analysis and Machine Intelligence (Top Picks for
30 years). In 2017 Google Scholar listed his work on segmentation
as a “classic paper in computer vision and pattern recognition” (from
2006). In 2011, he received the Helmholtz Prize from the IEEE and
the Test of Time Award from the International Conference on Computer Vision
\end{IEEEbiography}

\end{document}

\title{Occlusion-Ordered Semantic Instance Segmentation: Supplementary Materials}

\author{{ Soroosh Baselizadeh, Cheuk-To Yu, Olga Veksler}
        and~Yuri~Boykov
%\IEEEcompsocitemizethanks{\IEEEcompsocthanksitem  This work was performed while S. Baselizadeh was with the School of Computer Science, University of Waterloo. CT Yu, O. Veksler and Y. Boykov are with the School of Computer Science, University of Waterloo, Canada.\protect\\
%\IEEEcompsocthanksitem J. Doe and J. Doe are with Anonymous University.
}
% <-this % stops an unwanted space

% Soroosh commented the below two lines for arXiv
% \markboth{Journal of \LaTeX\ Class Files,~Vol.~14, No.~8, August~2015}%
% {Shell \MakeLowercase{\textit{et al.}}: Bare Demo of IEEEtran.cls for Computer Society Journals}

\maketitle

\IEEEdisplaynontitleabstractindextext
\IEEEpeerreviewmaketitle

\section{Details of Joint Semantic Segmentation and Oriented Occlusion Boundary Estimation}
\label{sec:joint_sem_edge_supp}

In this section, we provide more details about our deep model for joint semantic segmentation and oriented occlusion boundaries, including the probabilistic interpretation of our loss function in Eq.~(\ref{eq:our_3loss}) of the main paper.

\subsection{Architecture}
\label{sec:architecture}

\begin{figure*}
  \centering
  \begin{subfigure}[b]{1\textwidth}
  \includegraphics[width=0.8\textwidth]{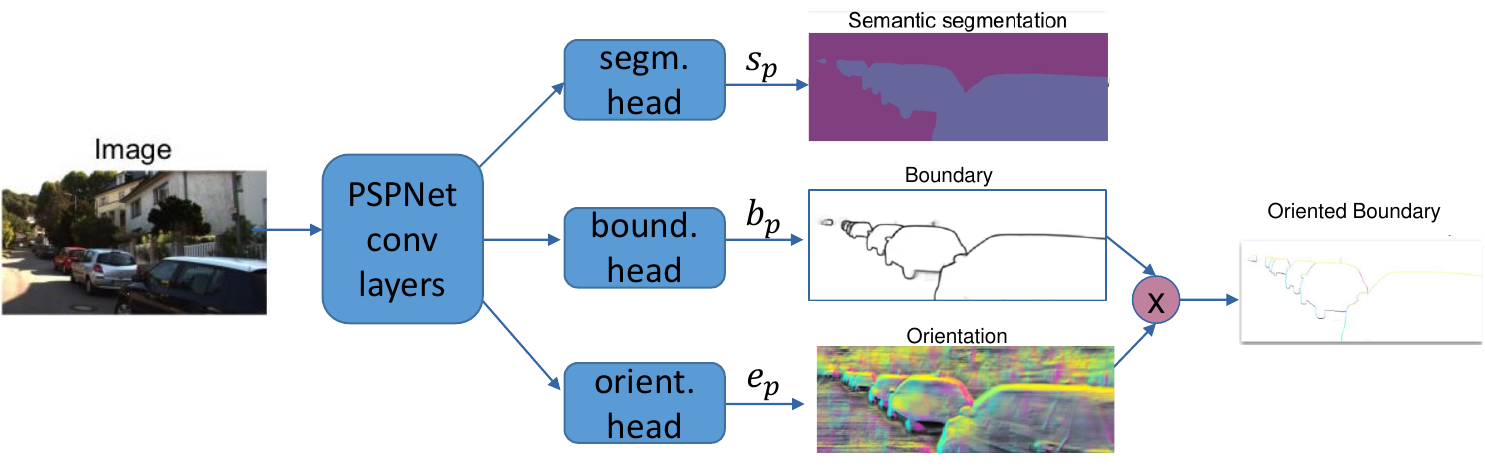}
  \caption{The overview of architecture for joint semantic segmentation and oriented occlusion boundary estimation.
  %We discard these approaches based on qualitative results due to poor performance.
  }
  \label{fig:edge-model-overview}
  \end{subfigure}
  \begin{subfigure}[b]{1\textwidth}
    \centering
    \includegraphics[width=.6\textwidth]{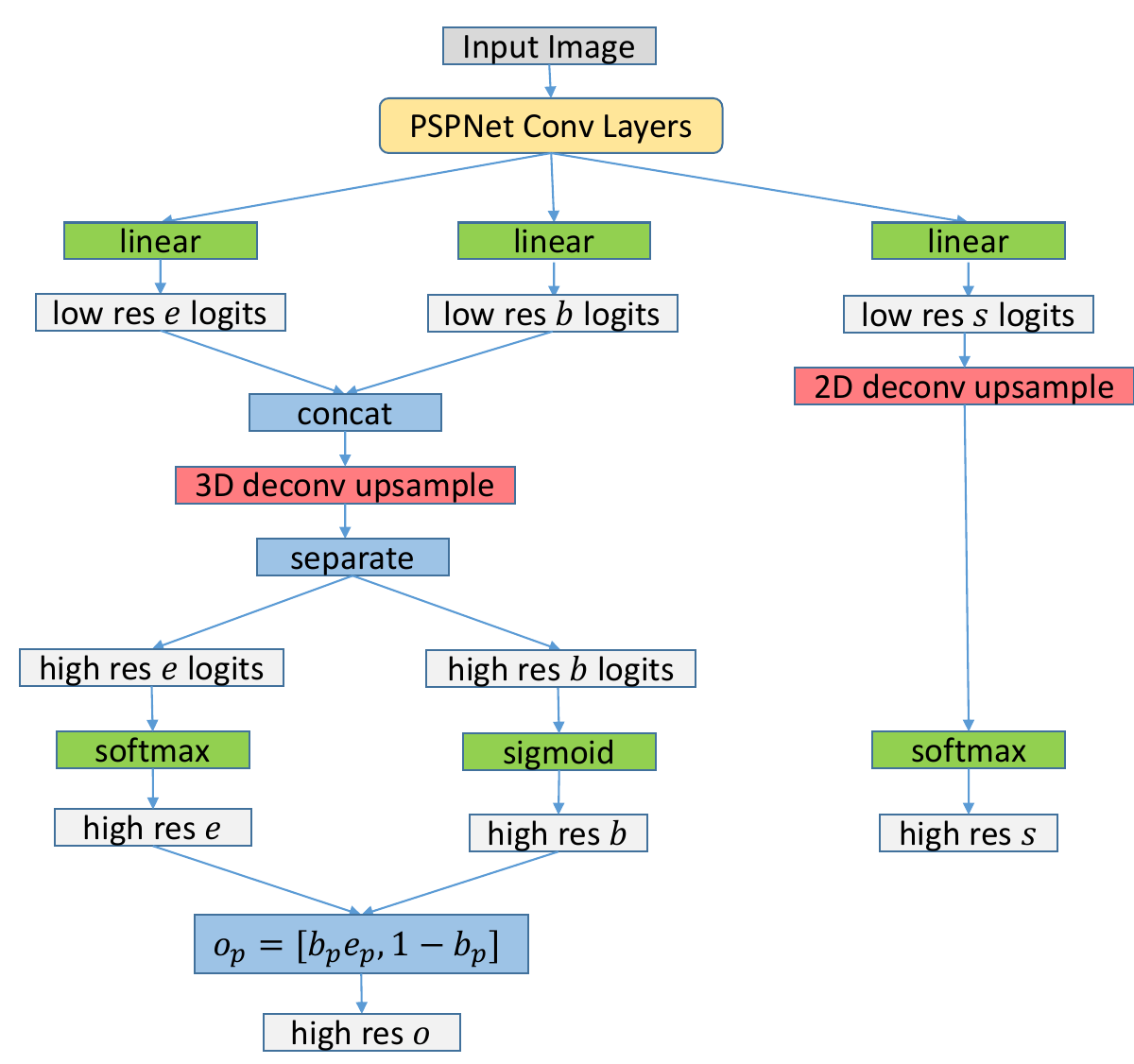}
    \caption{The detailed illustration of our deep model for joint semantic segmentation and oriented occlusion boundaries. See~\ref{sec:joint_sem_edge_supp} for the explaination.}
    \label{fig:edge-model}
    \end{subfigure}
    \caption{Visualization of our joint semantic segmentation and oriented occlusion boundary model.}
    
   \label{fig:joint_sem_edge_arch}
\end{figure*}

The overview of our architecture with three heads is in Fig.~\ref{fig:edge-model-overview}.
To obtain the two final predictions, our PSPNet-based model incorporates three interconnected heads:
\begin{enumerate}
\item  Head $s$: estimates of $Pr(\mathbb{S}_p)$ for each pixel $p$. This is similar to a standard semantic segmentation task. 
\item
 Head $b$: estimates $Pr(\mathbb{B}_p=1)$, the probability that pixel $p$ is an occlusion boundary. 
\item
 Head $e$: estimates the conditional distribution $Pr(\mathbb{O}_p=d | \mathbb{B}_p=1)$, which is the distribution of the normal at $p$ over $|D|$ bins, assuming $p$ is a boundary. %This probability has $D$ values similar to what we explained for $Pr(\mathbb{O}_p, \mathbb{B}_p=1)$. 
\end{enumerate}

We now provide a detailed explanation of the model stages, see~ Fig.\ref{fig:edge-model}.

\textbf{Features:} 
The input image is processed by PSPNet \cite{zhao2017pspnet} to extract deep features from the image.

\textbf{Logits:} Three separate linear layers are employed, each producing logits for the corresponding heads described above. As the output of PSPNet is of low resolution, we  upsample the logits in the subsequent stage.

\textbf{Upsampling:} For the $s$ logits, we utilize 2D (spatial) deconvolution upsampling. The upsampling is applied separately for each channel (each semantic class), as denoted by the "2D" descriptor. We initialize the upsampling kernel with bilinear upsampling weights. On the other hand, for upsampling the $b$ and $e$ logits, which contain complementary information, we concatenate them and perform 3D deconvolution upsampling. The "3D" nature of the kernel indicates that the channels (representing different orientation bins and $b$) effectively utilize each other's information. Finally, we separate the upsampled $b$ and $e$ logits.

\textbf{Heads:} Non-linearities are applied to the logits at this stage. Specifically, softmax is used for the $s$ and $e$ logits, while sigmoid activation is employed for the $b$ logits.

% \textbf{Final Outputs:} For the $s$ head, no further processing is required. It directly estimates the desired probability distribution $Pr(\mathbb{S}_p)$. For the oriented occlusion boundaries, we need to obtain {\em oriented boundary} prediction $o_p:=Pr(\mathbb{O}_p)$, 
% which is a distribution over $|D|+1$ values. 
% The basic probability relations in~\cref{eq:O=empty,eq:O=d} imply
% a simple formal relation of this prediction to $b_p = \Pr(\mathbb{B}_p=1)$ and to
% conditional orientation distribution $e_p=\Pr(\mathbb{O}_p=d | \mathbb{B}_p=1)$ 
% defined in the list above. 
% Indeed, separating $|D|+1$ components of vector $o_p$ into two parts: $|D|$ values
% corresponding to the probabilities $\Pr(\mathbb{O}_p=d)$ for $d\in D$ and 
% an extra value corresponding to $\Pr(\mathbb{O}_p =\emptyset)$, equations \eqref{eq:O=empty},\eqref{eq:O=d} give
% \begin{equation}
%     o_p := [b_pe_p, 1-b_p].
% \end{equation}
%, providing the joint probability distribution of occlusion boundary presence and its corresponding normal orientation. We call this final output as ``o''.

\subsection{Probabilistic Interpretation}
\label{sec:prob_interpretation}

\textbf{Loss:} We denote the ground truth for $s_p$, $b_p$, and $e_p$ by $S_p$, $B_p$, and $E_p$, respectively. Additionally, the final output of our model is $o$. Let $O_p$ indicate the ground truth for it.
%It is obvious that the sum of $O_p$ coordinates is $B_p$. That is $|O_p|_1=B_p$. We have the same thing for $|o_p|_1 = b_p$. This notion of $O_p$ and $o_p$ is representing the joint probability $Pr(\mathbb{E}_p, \mathbb{B}_p=1)$. This is what we are interested in as the probability of $p$ not being a boundary can be obtained by $1-b_p = 1-|o_p|_1$ or for ground truth $1-B_p = 1-|O_p|_1$. 

%Now, assume $[O_p,1-B_p]$ is the concatenation of $O_p$ and $1-B_p$ and $[o_p, 1-b_p]$ is the concatenation of $o_p$ and $1-b_p$. We know these newly defined vectors represent a distribution and sum to 1 as well. The last coordinate of them shows the probability of $p$ not being a boundary and the other ones show the joint probability of $p$ being a boundary and having one of the $D$ possible orientations. Hence, each of these new vectors has a $(D+1)$ dimension.
\begin{theorem}
Denoting cross-entropy by $CE$, we prove the following:
\begin{equation}
    \sum_p CE(O_p | o_p) = \sum_p CE(B_p|b_p) + B_p CE(E_p|e_p)
\end{equation}

\begin{proof}
Following the definition of $\mathbb{O}_p$ in Sec.~\ref{sec:joint_sem_edge} of the main paper, we have
\begin{equation}
    \label{eq:loss-proof}
    \begin{split}
        \sum_p CE(O_p | o_p) &= \sum_p -(1-B_p) \ln(1-b_p) \\
        &+ \sum_p \sum_{d \in D } -B_pE_p^d \ln{b_pe_p^d}
    \end{split}
\end{equation}

Now, consider only the last term:
\begin{equation}
    \begin{split}
        \sum_{d \in D } -B_pE_p^d \ln{b_pe_p^d} 
        &= \sum_p B_p \sum_{d \in D } -E_p^d (\ln{b_p} + \ln{e_p^d}) \\
        &= \sum_p B_p \sum_{d \in D } -E_p^d \ln{b_p} -E_p^d \ln{e_p^d} \\
        &= \sum_p B_p \sum_{d \in D } -E_p^d \ln{b_p} \\
        &+ \sum_p B_p \sum_{d \in D } -E_p^d \ln{e_p^d} \\
        &= \sum_p -B_p \ln{b_p} \sum_{d \in D } E_p^d \\
        &+  \sum_p B_p \sum_{d \in D } -E_p^d \ln{e_p^d}
    \end{split}
\end{equation}

Now, as we know $E_p$ is a conditional distribution, and it sums to 1, we have:
\begin{equation}
    \begin{split}
    \sum_{d \in D } -B_pE_p^d \ln{b_pe_p^d} &= \sum_p -B_p \ln{b_p} +  \sum_p B_p \sum_{d \in D } -E_p^d \ln{e_p^d}
    \end{split}
\end{equation}

Now, we substitute this back in Eq.~(\ref{eq:loss-proof})
\begin{equation}
    \begin{split}
        \sum_p CE(O_p | o_p) &= \sum_p -(1-B_p) \ln(1-b_p) + \sum_p -B_p \ln{b_p} + \\
        &\quad \sum_p B_p \sum_{d \in D \ bins} -E_p^d \ln{e_p^d} \\
        &= \sum_p CE(B_p|b_p) + \sum_p B_p CE(E_p | e_p) \\
        &= \sum_p CE(B_p|b_p) + B_p CE(E_p|e_p)  %\qedhere 
    \end{split}
\end{equation}
%\qed
\end{proof}
\end{theorem}

This suggests that instead of applying cross-entropy ($CE$) loss directly on the predictions $o$, we can apply it separately on the predictions $b$ and $e$. However, it is important to note that the cross-entropy loss for $e$ should only be computed for the ground-truth boundaries. This distinction aligns with the fact that the ground truth $O_p$ is defined for all pixels, whereas $E_p$ is only defined for ground-truth boundaries. As discussed in Sec.~\ref{sec:joint_sem_edge}, main paper, where we address the imbalance present in $b$, we utilize weighted cross-entropy ($wCE$) loss for the $b$ head, as it has been shown to provide improved performance in such cases. For the $s$ head, we employ standard cross-entropy loss, as explained in Sec.~\ref{sec:joint_sem_edge}, main paper. The final loss is also shown in Eq.~(\ref{eq:our_3loss}) of the main paper.

\subsection{Acquiring Ground Truth Boundaries} 
\label{sec:acquiring_gt_boundries}

For KINS and COCOA datasets, we use $|D|=4$. Having the instance masks, one can easily get the boundary pixels. Hence, $B_p$ is easy to produce. Regarding $E_p$, which is only defined for ground-truth boundaries, we determine the orientation distribution for each boundary pixel $p$. To achieve this, we examine whether pixel $p$ occludes any of its left, right, top, or bottom neighbor pixels. We then set the corresponding components in the orientation vector to 1, indicating the presence of occlusion in those orientations. Finally, to ensure that the vector represents a valid probability distribution, we normalize it to have a sum of 1. For example, if pixel $p$ only occludes its left neighbor, the corresponding $E_p$ vector is $[1, 0, 0, 0]$. On the other hand, if pixel $p$ occludes both its left and top neighbors, the $E_p$ vector is $[0.5, 0, 0.5, 0]$, indicating that the normal's orientation lies between the two directions (i.e., 45 degrees to the top left). 

\subsection{Testing Oriented Occlusion Boundary}
\label{sec:testing_oob}
 Prior works~\cite{wang2016doc, wang2019doobnet, qiu2020pixel} compare their methods using the angle of boundary using POR curves. Hence, to compare, we also convert our quantized predictions to an angle.
To convert, we proceed as follows, keeping in mind that our orientation vector estimates are noisy. An orientation cannot be simultaneously to the left and to the right. Therefore, we take the maximum of the left and  right components from $o$. Similarly, orientation cannot be simultaneously to the top and bottom, therefore, we take  the maximum of the top and bottom components from  $o$'. We set the angle of the normal by finding the arc tangent of the achieved vector of the two mentioned maximums and their directions. For example, if $o_p=[0.5, 0.1, 0.4, 0.0]$ the angle is $\alpha = arctan{\frac{0.5}{0.4}}$.  Furthermore, we convert the angle of the normal to the angle of the boundary using the left rule \cite{wang2016doc}, which involves adding or subtracting $\frac{\pi}{2}$.

% %\section{Occlusion-Order based Instance Grouping: Optimization Details}
% %\label{sec:optimization}
% \section{Details of CRF Occlusion Order Labeling}
% \label{sec:instance_occlusion_labeling_supp}
% In Sec.~\ref{sec:instance_occlusion_labeling} of the main paper we describe the second step of our OOSIS approach, which involves  a formulation and optimization of an energy function 
% in Sec.~\ref{eq:main-energy}.
% We now give more details about our approach for minimizing this energy as well as other energy minimization-related discussion.

\section{The Expansion vs. Jump move Algorithms}
For minimizing CRF energies of the type of Eq.~(\ref{eq:main-energy}) of the main paper, the expansion algorithm~\cite{BVZ:PAMI01} is frequently used. The expansion move has approximation guarantees and has been shown superior to other minimization algorithms for many energy types~\cite{MRFStudy:ECCV06}. However, we found that the expansion algorithm does not work as well for our energy as the jump move, and now explain the reasons.

The expansion algorithm performs a sequence of $\alpha$-expansion moves until convergence. Given the current labeling and some label $\alpha$, an $\alpha$-expansion move finds the subset of pixels to switch to label $\alpha$ s.t. the energy decreases by the largest amount. 

Typically, one starts with a labeling where all pixels are assigned label 0, and then performs a series of cycles, where each cycle consists of one iteration over labels in $\{0,1,2,...,l_{\max}\}$. Here $l_{max}$ is the largest possible label. Notice that iterations are performed not necessarily in the consecutive order of labels.  In general, the expansion moves for our energy are not submodular. However, 
it is straightforward to show that if we start with the labeling where each pixel is assigned $0$, and then expand on labels $1,2,..., l_{\max}$, in that order, then each expansion is submodular, i.e. one cycle of the expansion algorithm is submodular if labels we expand on are in the increasing order. If we perform more than one cycle, then the expansions are not necessarily submodular. 

Let $\bx^0$ be a lafbeling where each pixel is assigned 0. 
Consider the example in Fig.~\ref{fig:jumps} in the main paper.
Suppose we apply one cycle of expansions, in the increasing order, starting with $\bx^0$.
Let us refer to the car objects as car 1, 2, 3, in the order of their depth, with  car 3 being the front-most. 
After the first expansion, i.e. expansion on label $1$, we will get the same result as in Fig.~\ref{fig:jumps}c, with all cars labeled with  1. This is because if we start with $\bx^0$, both the expansion move on label 1 and the jump move are submodular and can be optimized exactly, and, furthermore, the optimal jump and expansion moves coincide. The next expansion, i.e. expansion on label 2, will produce the labeling as in Fig.~\ref{fig:jumps}d and is the same as the second jump move, where car 3 gets label 2, and cars 1 and 2 stay with label 1. Switching cars 2 and 3 to label 2 would result in a worse energy (jump move would switch cars 2 and 3 to label 2 if that was lower energy than the energy 
in Fig.~\ref{fig:jumps}d. 
Any further expansion moves will not change the labeling, unlike further jump moves. To get to the lower energy labeling in Fig.~\ref{fig:jumps}e, we need to simultaneously switch car 2 to label 2 and switch car 3 to label 3. Jump move is able to do this, but the expansion algorithm cannot switch pixels to two different labels.

\section{Additional Experiments}
\label{sec:additiona_experiments_suppl}

\subsection{Qualitative Examples}
\begin{figure}
    \centering
    \includegraphics[width=0.5\textwidth]{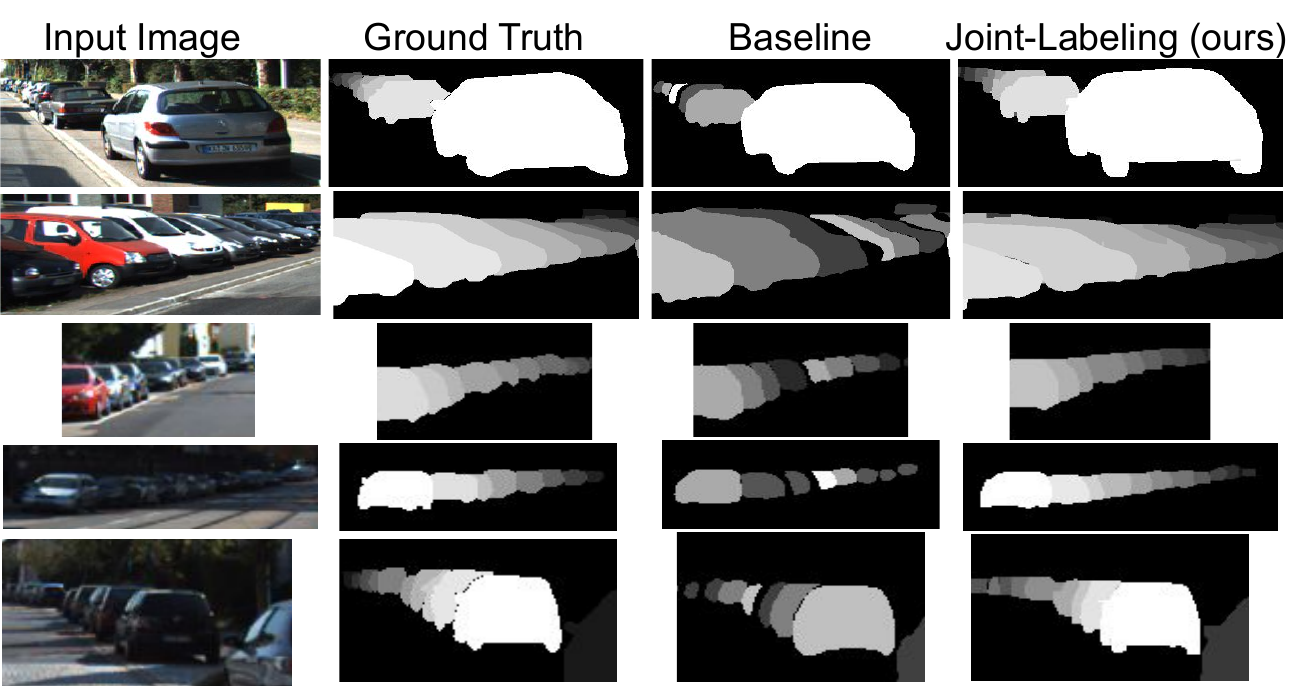}
    \caption{Relative depth map visualization of OOSIS for baseline and our approach on the KINS dataset. The best version of the baseline is displayed. The images are cropped and zoomed in for better visualization of differences.}
    \label{fig:qual-supp}
\end{figure}
\begin{figure}
    \centering
    \includegraphics[width=0.5\textwidth]{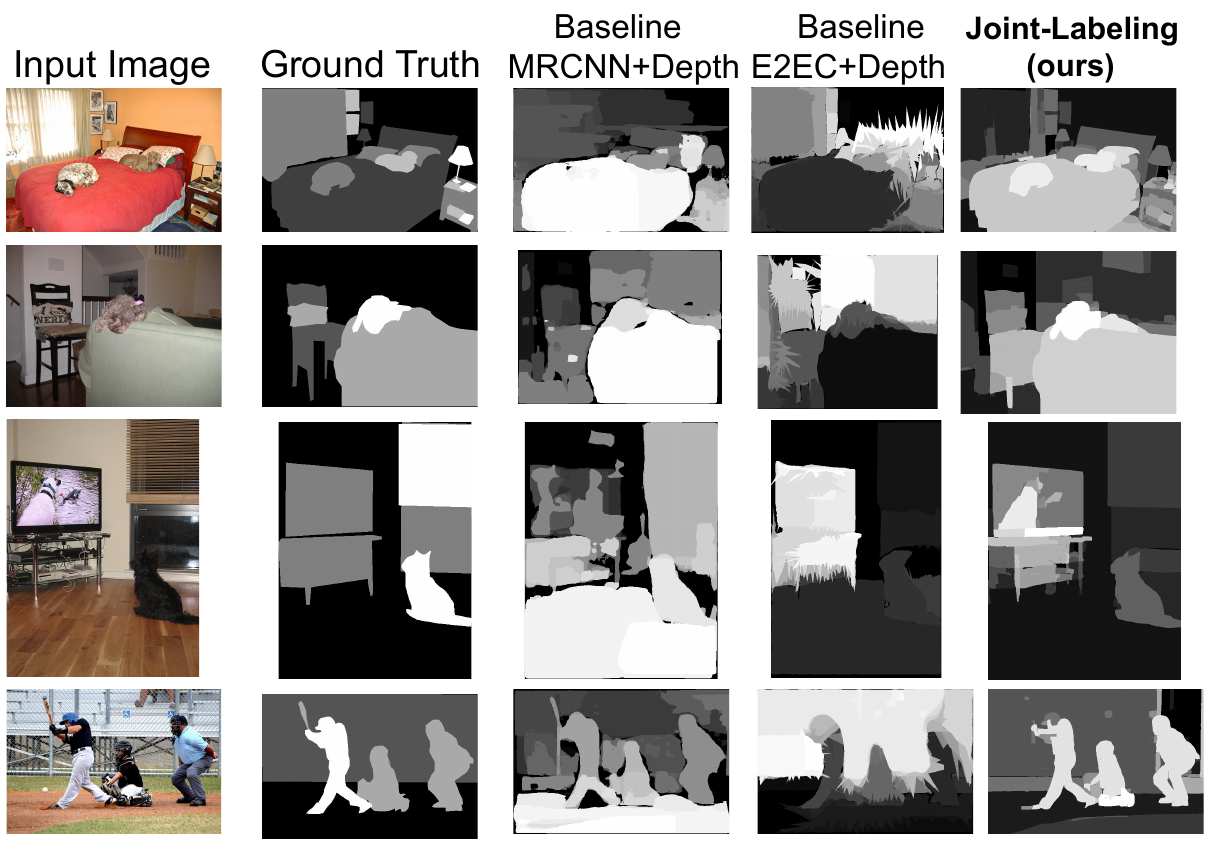}
    \caption{Relative depth map visualization of OOSIS for baseline and our approach on the COCOA dataset. The best versions of the baselines are displayed.}
    \label{fig:qual-supp-cocoa}
\end{figure}
\begin{figure*}
    \centering
    \includegraphics[width=1\textwidth]{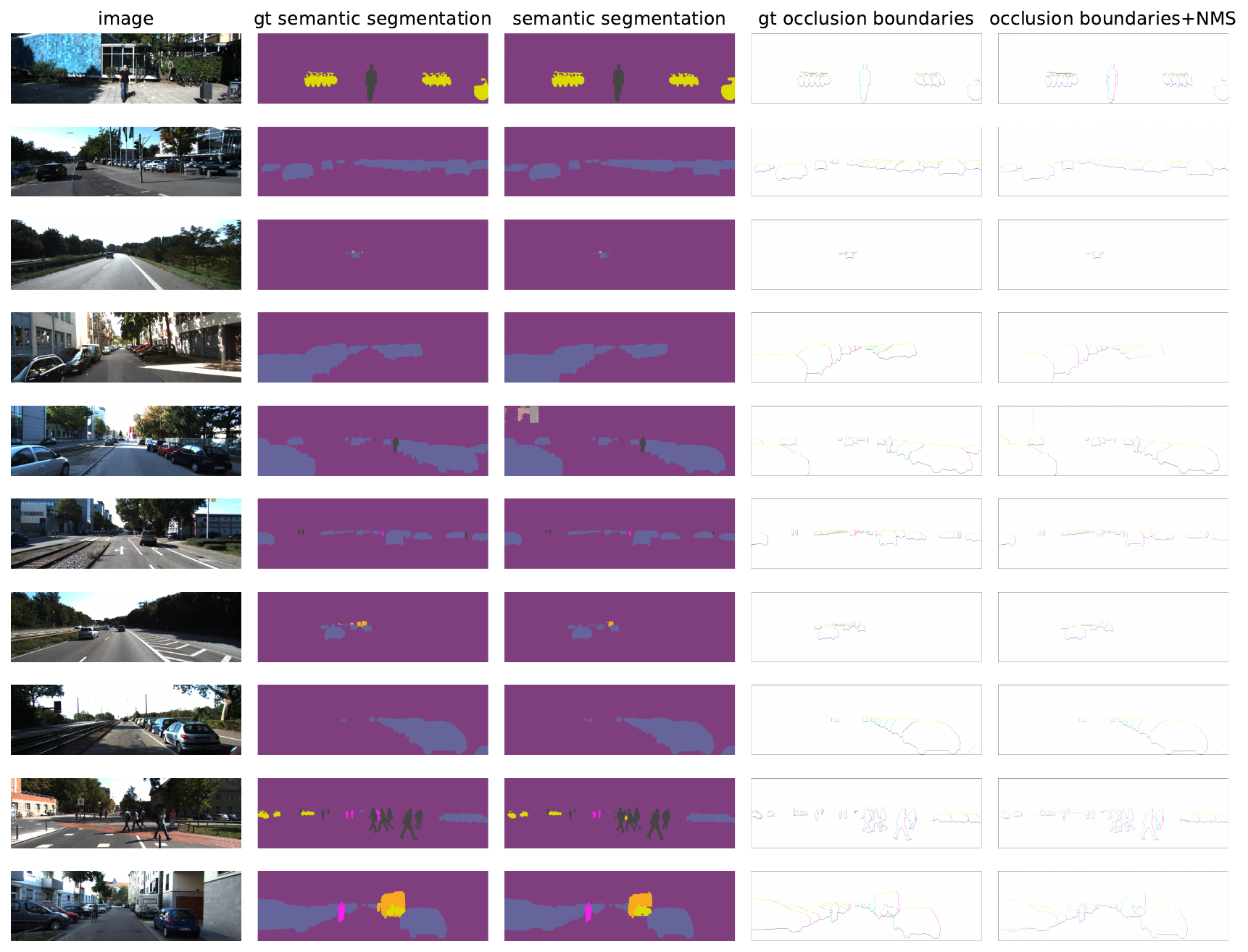}
    \caption{Illustration of our semantic segmentation and oriented occlusion boundaries  for our model. The images are: input, ground truth and our semantic segmentation, ground truth and our oriented occlusion boundaries after non-maximum suppression. The color scheme for the boundaries is: left-cyan, top-yellow, right-magenta, bottom-black. Best viewed zoomed in.}
    \label{fig:boundaries-supp}
\end{figure*}
Fig.~\ref{fig:qual-supp} and Fig.~\ref{fig:qual-supp-cocoa} show visualizations of our approach vs. the best version of baseline and the ground truth for more images of KINS and COCOA respectively. Fig.~\ref{fig:boundaries-supp} represents  the output of our deep model for semantic segmentation and oriented occlusion boundary for more examples of KINS.

\subsection{InstaOrder Dataset}
\begin{figure}
    \centering
    \includegraphics[width=0.3\textwidth]{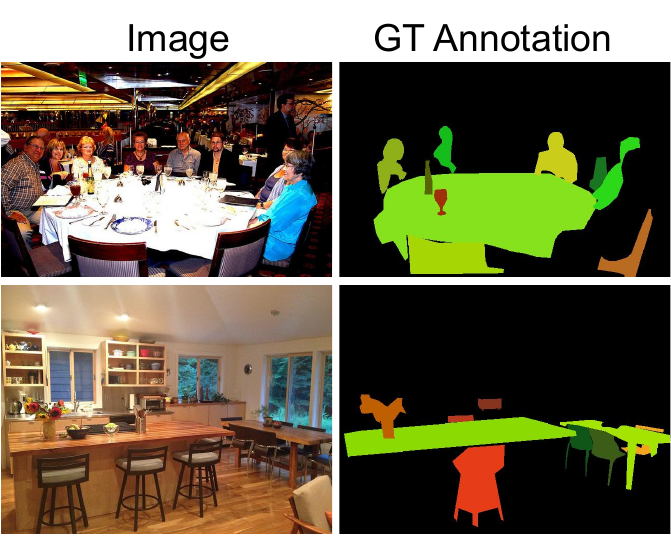}
    \caption{Examples of the annotations in the InstaOrder dataset~\cite{lee2022instance}. Only a fraction of objects in a scene are annotated, which confuses segmentation models for our task of OOSIS.}
    \label{fig:instaorder_examples}
\end{figure}
The authors in~\cite{lee2022instance} propose a new dataset, called InstaOrder dataset, for the task of pairwise occlusion ordering of \emph{``provided''} instances. This dataset, characterized by a considerable volume of images, may initially appear conducive to accommodating the OOSIS task. However, as stated in their work, due to resource constraints in human annotation efforts, only a maximum of 10 instances per image received annotations. While this limited instance annotation count holds potential for ordering predetermined instances effectively, it significantly hampers the dataset's suitability for OOSIS—a task involving both segmenting instances and determining their order. The primary issue arises from the dataset's nature: it contains numerous objects, potentially belonging to the same class, but only a subset of the objects is annotated. 
The sparsity of annotation of InstaOrder dataset~\cite{lee2022instance}  can be observed in Fig.~\ref{fig:instaorder_examples}, where multiple instances of humans in the upper image and chairs in the lower row remain not annotated despite their presence.

When training our joint oriented occlusion boundary and semantic segmentation CNN, every pixel must be annotated. If we annotate everything which is not a part of an annotated instance as the background, the resulting  'ground truth' would contain large erroneously marked regions, and the trained model would be of an unacceptable accuracy. If we annotate everything which is not part of an annotated instance as 'void', then we would not have any background samples, and the resulting model would be of even lower accuracy.
%Consequently, the model encounters ambiguity regarding whether to extract unannotated instances, leading to confusion in the segmentation process.

This situation is even worse for model evaluation, where instances correctly predicted by a model but absent in the annotated ground truth would be treated as false positives. Consequently, the dataset fails to facilitate fair model evaluations, skewing results due to discrepancies between model-generated instances and the limited ground truth annotations.

 Since the images in InstaOrder dataset are a subset of the COCO dataset, one might think it is feasible to use the annotations of COCO instead for instances. However, COCO annotations do not have occlusion order annotation, and hence, cannot be used for OOSIS as well. Thus  we do not use InstaOrder dataset in this work for OOSIS task.

\section{Implementation Details}
%\label{sec:code}
%\subsection{Implementation Details} 
\textbf{For KINS:} For E2EC, we use their official implementation and configurations, and we only train it on modal masks. Hence, the model is trained for 150 epochs. For Mask R-CNN, we use the implementation of~\cite{tran2022aisformer}. For InstaOrder and OrderNet, we use the pre-trained models of~\cite{lee2022instance}. For PanopticMask2Former~\cite{cheng2021mask2former} we use the implementation in~\cite{wu2019detectron2} with a ResNet50 backbone. The default configurations for Cityscapes dataset was sued for KINS due to the similarity of driving scenes with the image size-related configurations adapted to fit the size and aspect ratio of the KINS dataset. The number of training epochs was also adjusted to 80 for KINS. For our joint semantic segmentation and oriented occlusion boundary model, we use the PSPNet~\cite{zhao2017pspnet} from the implementation of~\cite{semseg2019}. We use weight=$0.9$ with $wCE$ in our deep model for semantic segmentation and occlusion boundary and train for 200 epochs using SGD and weight decay as in the implementation, we used a single GPU and a batch size of 8. We train using crop size of 225x225 and test on a resolution of 2048x615. The testing resolution is similar for all models. The backbone of our PSPNet is Resnet-50, similar to the backbone of Mask R-CNN that we used. P2ORM~\cite{qiu2020pixel} is also used based on their official implementation and the settings were used as in their code or paper. For a fair comparison, we used the 4-neighborhood variant of theirs as we use a 4-neighborhood system in our deep model. For depth, we use~\cite{lyu2021hr}, trained on KITTI dataset, from their official repository.

\textbf{For COCOA:} For E2EC, we train for 50 epochs using Adam with parameters of their implementation for COCO and finetune for 10 epochs using SGD for maximum performance. For Mask R-CNN we use the implementation of \cite{mmdetection} and use their configuration for COCO with the 1x training regiment. Again, for InstaOrder and OrderNet, we use the pre-trained models from~\cite{lee2022instance}. For PanopticMask2Former~\cite{cheng2021mask2former} we use the implementation in~\cite{wu2019detectron2} with a ResNet50 backbone. The default configurations for COCO dataset was used for COCOA due to the similarity of data, particularly COCO being the mother dataset of COCOA. The number of training epochs was also adjusted to 50 for COCOA. Our setting for our deep model for semantic segmentation and occlusion boundary is similar to KINS. The test resolution for all models is the same as the original image size. The backbones are similar to KINS. 
For MiDaS~\cite{Ranftl2022}, we used the official implementation with ``midas\_v21\_384'' model type.

%\subsection{Code} We are attaching our code in the supplementary material. We will release the code publicly.

\bibliographystyle{IEEEtran}
\bibliography{eccv}